\documentclass[sigconf,nonacm]{acmart}
\AtBeginDocument{%
  }

\setcopyright{none}
\renewcommand\footnotetextcopyrightpermission[1]{}

\usepackage{float}
\usepackage{algorithm}
\usepackage{algpseudocode}
\usepackage{amsmath}
\usepackage{mathtools}
\usepackage{amsthm}
\usepackage{amsfonts}
\usepackage{nicefrac}
\usepackage{graphicx}
\usepackage{multirow}
\usepackage{makecell}
\usepackage[table]{xcolor}

\graphicspath{{media/}}

\begin{document}

\title{MolSight: A Graph-Aware Vision-Language Model for Unified Chemical Image Understanding}


\author{Wenda Wang}
\email{wangwenda87@ruc.edu.cn}
\orcid{0000-0003-2469-8522}
\affiliation{%
  \department{Gaoling School of Artificial Intelligence}
  \institution{Renmin University of China}
  \country{}
}
\affiliation{%
  \institution{Alibaba Group}
  \country{}
}

\author{Yihan Tong}
\email{2024201606@ruc.edu.cn}
\affiliation{%
  \department{Gaoling School of Artificial Intelligence}
  \institution{Renmin University of China}
  \country{}
}

\author{Yuwei Hu}
\email{huyuweiyisui@ruc.edu.cn}
\affiliation{%
  \department{Gaoling School of Artificial Intelligence}
  \institution{Renmin University of China}
  \country{}
}

\author{Xuchen Pan}
\email{panxuchen.pxc@alibaba-inc.com}
\affiliation{%
  \institution{Alibaba Group}
  \country{}
}

\author{Zhewei Wei}
\email{zhewei@ruc.edu.cn}
\orcid{0000-0003-3620-5086}
\affiliation{%
  \department{Gaoling School of Artificial Intelligence}
  \institution{Renmin University of China}
  \country{}
  \thanks{Zhewei Wei is the corresponding author.}
}

\author{Yaliang Li}
\orcid{0000-0002-4204-6096}
\email{yaliang.li@alibaba-inc.com}
\affiliation{%
  \institution{Alibaba Group}
  \country{}
}

\author{Bolin Ding}
\orcid{0000-0003-1535-9692}
\email{bolin.ding@alibaba-inc.com}
\affiliation{%
  \institution{Alibaba Group}
  \country{}
}

\renewcommand{\shortauthors}{Wang et al.}

\begin{abstract}
    Using molecular large language models (LLMs) as a unified framework for understanding molecular structures and functions is emerging as a new trend in tasks such as molecular design and drug discovery. However, these models struggle to fully capture the visual representation of molecular structures, limiting their potential. While existing molecular vision-language models (VLMs) show promise, they still face challenges in structural alignment and lack the necessary topological modeling for accurate molecular understanding. To address this, we propose MolSight, a graph-aware vision-language model framework designed to enhance the understanding of molecular images by VLMs. MolSight integrates a Molecular Topology Module to inject chemical-bond adjacency information into vision tokens, and a Molecular Grounding Module to align visual features with chemical symbolic semantics. Our experiments demonstrate that MolSight significantly outperforms existing VLMs, molecular LLMs, and task-specific models across multiple chemical visual understanding tasks, achieving a new level of molecular image reasoning in complex chemical scenarios.
\end{abstract}




\maketitle

\section{Introduction}\label{introduction}

Accurately identifying molecular structures and inferring their physicochemical properties and biological functions are fundamental to molecular design and drug discovery~\citep{stokes2020deep,merchant2023scaling}. This understanding process is inherently multi-modal: researchers usually need to combine molecular structure images, SMILES strings, and natural-language descriptions to identify key structural features and reason about molecular properties and functions~\citep{weininger1988smiles,edwards2022translation}. By learning general chemical knowledge from large-scale molecular data, large language models can perform molecular generation, optimization, and property prediction within a unified framework~\citep{edwards2022translation,pei2023biot5,zhang2024chemllm}, thereby promoting the transition of molecular reasoning and design from domain-specific methods toward a unified “AI chemist” paradigm across diverse chemical reasoning scenarios~\citep{bran2023chemcrow,boiko2023autonomous}.

Compared with generalist LLMs, Molecular LLMs need the key ability to accept molecular languages that contain chemical structural information. The most common input format is canonical SMILES, which contains the corresponding molecular graph structure~\citep{weininger1988smiles}: representative molecular LLMs, such as MolT5~\citep{edwards2022translation}, BioT5 \citep{pei2023biot5}, MolCA~\citep{liu2023molca}, and ChemLLM~\citep{zhang2024chemllm}, have achieved promising progress on molecular understanding tasks. However, this paradigm has two limitations. First, LLMs directly receive SMILES as linear text sequences, while molecular graph topology is implicitly encoded in SMILES. Although these two forms should represent equivalent structural information, current LLMs still suffer from clear information loss when perceiving molecular structural semantics directly from SMILES in practice. Meanwhile, this paradigm is also inconsistent with the actual multi-modal workflow of chemical researchers, where molecular structure images remain an important medium for recording and reading chemical structures~\citep{krasnov2024comparing,morin2024patcid}.

Vision-language models (VLMs) can directly recognize molecular images, and therefore logically provide a more natural interface for injecting molecular graph structural semantics~\citep{molvision,li2025chemvlm,tan2025chemmllm}. However, both prior studies and our preliminary experiments show that current VLMs have severely insufficient molecular understanding ability. On the MolVision~\citep{molvision} bioactivity prediction task, image-based inputs significantly underperform SMILES-based inputs. In the image-to-SMILES translation task, generalist VLMs achieve near-zero accuracy. Although multi-modal models trained with chemical image adaptation obtain some improvements, their structural recognition ability still significantly lags behind specialized optical chemical structure recognition (OCSR) tools such as MolScribe~\citep{molscribe} and DECIMER~\citep{decimer}. These results indicate that current VLMs have not truly learned the structural semantics in molecular images.

\begin{figure}[t]
    \centering
    \includegraphics[width=\columnwidth]{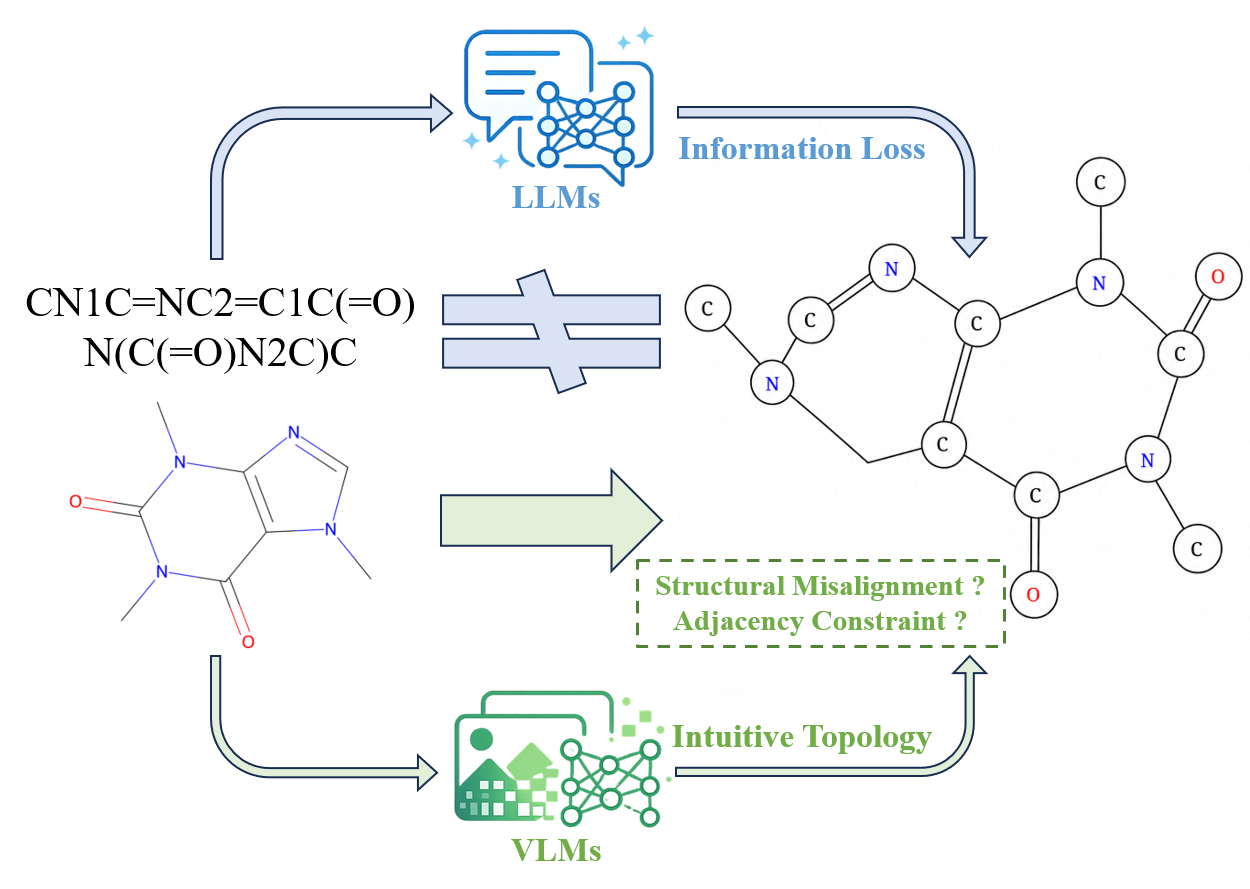}
    \caption{%
    \textbf{MolSight's motivation illustration.}
    Linear SMILES sequences are structurally equivalent to molecular graphs, yet current LLMs still suffer from graph-topological information loss when processing such linear representations. Molecular images intuitively preserve atoms, bonds, and connectivity, but existing VLMs still face structure-semantic misalignment and insufficient adjacency modeling.}
    \Description{motivation}
    \label{fig:motivation}
\end{figure}

We argue that this failure originates from the structural misalignment between visual representations produced by standard vision encoders~\citep{dosovitskiy2020image,radford2021learning,li2023blip2,liu2023visual} and molecular structural semantics. Vision encoders pretrained on general images mainly capture patterns such as color, texture, and spatial layout~\citep{dosovitskiy2020image,radford2021learning,caron2021emerging}, which are not suitable for describing sparse and fine-grained atom and chemical bond features. In addition, the vision tokens passed to the LLM mainly preserve locality in image space rather than adjacency relationships in molecular graphs \citep{kipf2016semi,gilmer2017neural,velivckovic2017graph}, and they lack explicit graph-structural constraints. As a result, it is difficult to preserve molecular topology information at the representation level, such as substructures including aromaticity and ring systems ~\citep{battaglia2018relational,liu2023molca}.

Motivated by the above progress and bottlenecks, we propose \textbf{MolSight}, a graph topology-aware vision-language framework for molecular image understanding. MolSight takes the molecular image and its image-derived SVG representation as inputs, where the SVG serves as an image-side structural annotation rather than an independent molecular sequence or external graph annotation. MolSight then introduces two complementary modules between the vision encoder and the LLM decoder: the \textbf{Molecular Topology Module (MTM)} and the \textbf{Molecular Grounding Module (MGM)}. The former injects molecular graph structure into vision tokens through learnable topology modeling, while the latter aligns vision tokens with symbolic annotations in the SVG through cross-modal attention. Together, they form a graph-aware topology adaptation layer that enables a general-purpose VLM backbone to reliably capture the chemical semantics in molecular images. Our main contributions are as follows:
\begin{itemize}
    \item To address structural misalignment between VLM visual representations and molecular image semantics, we use the image-derived SVG as an image-side structural cue. It preserves atom symbols, bond primitives, and 2D positions from the same image, rather than introducing independent molecular sequences or external graph annotations.

    \item We design two complementary modules: the \textbf{Molecular Topology Module (MTM)} explicitly injects molecular topology into vision tokens through a learnable edge predictor, and the \textbf{Molecular Grounding Module (MGM)} aligns vision tokens with the image-derived SVG representation via cross-attention. Together, they jointly form the graph topology adaptation layer of MolSight.

    \item Across four different types of chemical visual understanding tasks, MolSight outperforms existing generalist VLMs and molecular specialist VLMs, demonstrating its ability as a unified multi-modal molecular understanding model.
\end{itemize}

\begin{figure*}[!ht]
    \centering
    \includegraphics[width=\textwidth]{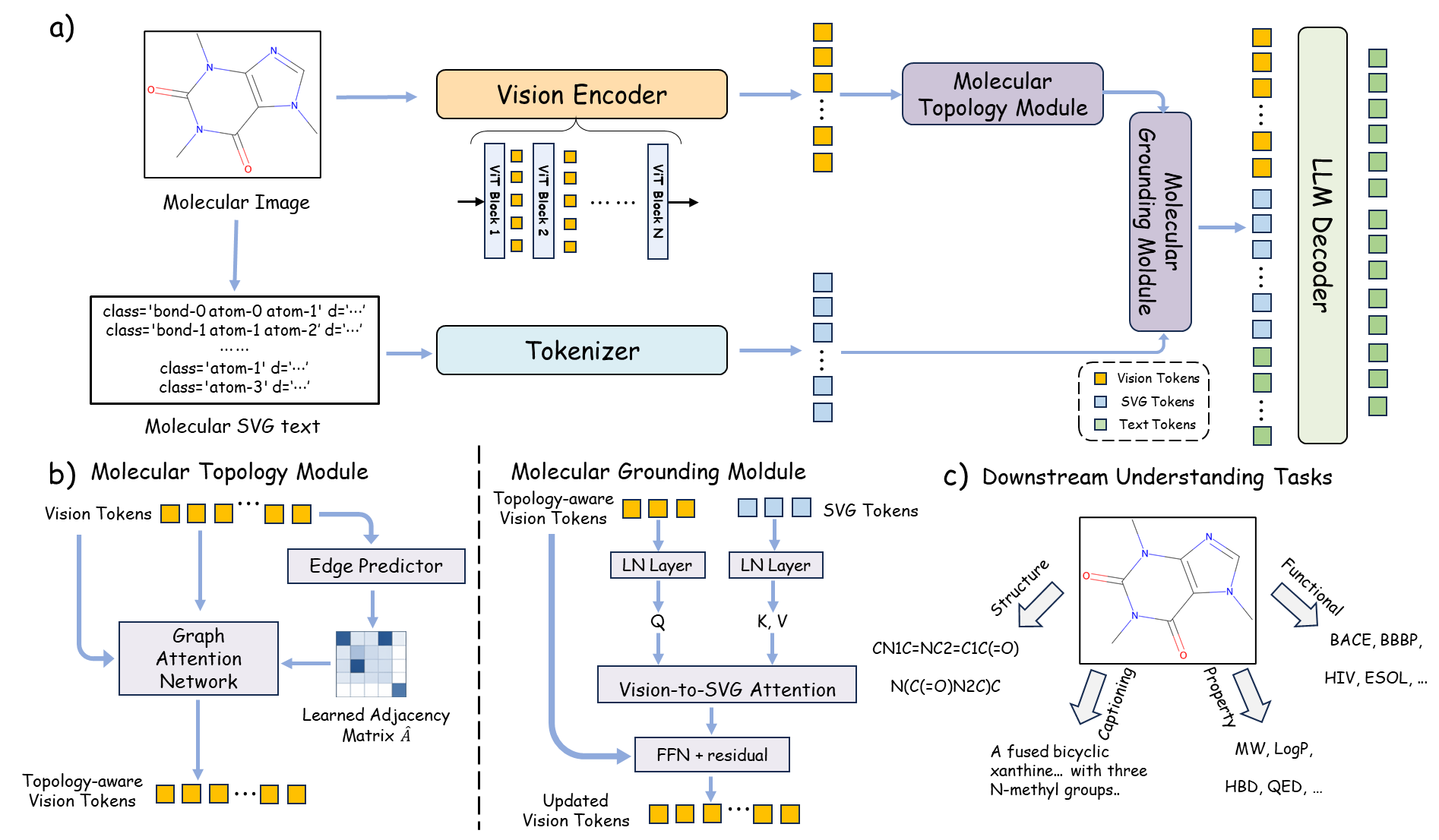}
    \caption{%
    \textbf{Overview of MolSight.}
    \textbf{a)} Overall framework. Molecular image tokens and SVG text tokens are updated and integrated through molecular topology and grounding modules before LLM decoding.
    \textbf{b)} The Molecular Topology Module injects molecular topology into vision tokens, while the Molecular Grounding Module aligns vision tokens with SVG text annotations, jointly introducing graph topology messages into vision-token representations.
    \textbf{c)} Four types of molecular visual understanding downstream tasks: SMILES translation, molecular captioning, descriptor estimation, and bioactivity prediction.
    }
    \Description{F1}
    \label{fig:overview}
\end{figure*}
\section{Related Work}\label{related_work}
\subsection{Multi-modal Molecular Language Models}
Molecular LMs aim to build general-purpose models capable of understanding and generating molecules across diverse tasks by learning molecular structures and chemical knowledge. Methods such as MolT5~\cite{edwards2022translation}, BioT5~\cite{pei2023biot5}, and ChemLLM~\cite{zhang2024chemllm} mainly inject chemical semantic knowledge into models by learning from large-scale chemical corpora and using SMILES as the primary molecular representation. Subsequent works, including MolCA~\cite{liu2023molca}, HIGHT~\cite{chen2024hight}, and Atomas~\cite{zhang2025atomas}, further explore multimodal molecular representation learning by aligning SMILES, molecular graphs, and natural-language descriptions, enabling models to connect different symbolic representations of the same molecule. GIT-Mol~\cite{liu2024gitmol}, ChemVLM~\cite{li2025chemvlm} and ChemMLLM~\cite{tan2025chemmllm} introduce the visual modality to leverage the explicit structural information provided by molecular images. In contrast, MolSight focuses on the underexplored graph-topological semantics in molecular images: by explicitly learning molecular topology from vision tokens and aligning it with structural annotations for topology-aware reasoning, MolSight improves the visual understanding and chemical semantic reasoning capabilities of VLMs for molecular images.

\subsection{Vision-language Models for Visual Reasoning}
Recently, vision-language models such as GPT-4V~\cite{achiam2023gpt4}, Flamingo~\cite{alayrac2022flamingo}, and Qwen-VL~\cite{bai2023qwen} have achieved remarkable progress on multimodal tasks, including visual question answering (VQA)~\cite{goyal2017making} and image captioning~\cite{lin2014microsoft}, by jointly learning visual and textual representations. As research attention has gradually shifted toward visual reasoning, benchmarks such as VisuLogic~\cite{xu2025visulogic} and VRB~\cite{nagar2024zero} show that existing VLMs still struggle with tasks that require logical reasoning based on precise spatial relations, performing close to random on several reasoning tasks. In addition, existing models often rely on external knowledge or language prompts during reasoning, rather than directly extracting relational and structural constraints from the visual modality itself~\cite{cui2023holistic}. In contrast, MolSight uses SVG annotations directly obtained from molecular images as auxiliary structural semantic signals, rather than relying on external knowledge, thereby enabling a more self-contained capture of molecular topology and chemical semantics from visual inputs.
\section{Methods}\label{methods}

MolSight aims to improve the visual understanding of molecular images by injecting graph-topological structural semantics into VLMs. Figure~\ref{fig:overview} provides an overview of MolSight, including its overall architecture, the two core topology adaptation components, and the supported chemical visual understanding tasks. We first introduce the \textbf{Molecular Topology Module (MTM)}, which predicts a chemical-bond adjacency matrix from vision tokens and uses it as a mask constraint to update visual representations. We then describe the \textbf{Molecular Grounding Module (MGM)}, which aligns visual features with chemical symbolic semantics through cross-modal attention between vision tokens and SVG text tokens. Then we present the model architecture and pipeline. Finally, we explain training strategy and loss functions.

\subsection{Molecular Topology Module}

The MTM aims to inject 2D molecular graph topology more explicitly into the vision tokens $V=\{v_1,\ldots,v_K\}\in\mathbb{R}^{K\times d}$. It first predicts chemical-bond connectivity between vision tokens through a learnable edge predictor, and then uses the predicted adjacency structure to guide graph-attention message passing.

For each vision token, we obtain its source and target representations through two low-dimensional projections for pairwise edge scoring, $h^{\mathrm{src}}_i = W_s v_i$ and $h^{\mathrm{dst}}_i = W_d v_i$. This design allows the edge predictor to learn a more expressive relation-specific compatibility function, while the following symmetrization preserves the undirected nature of chemical bonds. We compute the edge logit between vision tokens $i$ and $j$ as
\begin{equation}
    Z_{ij}
    =
    \frac{1}{2\sqrt{d_e}}
    \left[
    \langle h^{\mathrm{src}}_i, h^{\mathrm{dst}}_j \rangle
    +
    \langle h^{\mathrm{src}}_j, h^{\mathrm{dst}}_i \rangle
    \right],
    \qquad
    \hat{A}_{ij} = \sigma(Z_{ij}),
    \label{eq:mtm-edge}
\end{equation}
where $d_e$ is the dimension of the edge projection and $\sigma$ denotes the sigmoid function. The resulting soft adjacency matrix $\hat{A}$ encodes the model's estimate of chemical-bond connectivity between tokens. During training, the edge logits $Z$ are supervised by a token-level molecular adjacency matrix $A^*$ derived from the SVG representation. We parse the atom--bond relations and atom coordinates from the SVG, and map each atom to the corresponding cell in the merged visual-token grid according to its 2D coordinate. If two atoms are connected by a chemical bond, an edge is assigned between their corresponding visual tokens. We also add self-connections for tokens that contain atoms, so that atom-bearing visual regions are explicitly marked. Therefore, $A^*$ is not a raw atom-level adjacency matrix, but an SVG-derived token-level adjacency matrix aligned with the visual-token sequence, providing direct supervision for learning molecular topology from image tokens.

\textbf{Graph attention message passing.}
Given $\hat{A}$, we use it as a soft topological mask in multi-head attention:
\begin{equation}
    \alpha_{ij}
    = \mathrm{softmax}_j \left(
    \frac{\langle W_q v_i, W_k v_j\rangle}{\sqrt{d_h}}
    \cdot \hat{A}_{ij}
    -
    C(1-\hat{A}_{ij}) \right).
    \label{eq:mtm-attn}
\end{equation}
\begin{equation}
    v^{\mathrm{MTM}}_i
    =
    \mathrm{FFN}\!\left(
    v_i + W_o \sum_{j=1}^{K} \alpha_{ij} W_v v_j
    \right).
    \label{eq:mtm-update}
\end{equation}
where $d_h$ is the dimension of a single attention head, $C$ is a large positive constant that penalizes token pairs with low predicted adjacency, and $\mathrm{FFN}(\cdot)$ denotes the feed-forward transformation in the Transformer block. The resulting representation $V_{\mathrm{MTM}}=\{v^{\mathrm{MTM}}_i\}_{i=1}^{K}$ is topology-aware, as each token aggregates information according to the predicted molecular adjacency structure before being passed to the next module.

\subsection{Molecular Grounding Module}

The role of MGM is to align topology-aware vision tokens with the chemical symbolic annotations in the SVG text. After MTM injects molecular graph topology into the vision tokens, MGM allows each vision token to absorb atom types, bond connectivity, and positional information from the SVG through vision-to-SVG cross-attention.

\textbf{Vision-to-SVG cross attention.}
We take $V_{\mathrm{MTM}}$ as queries and the SVG text-token embeddings $E_{\mathrm{svg}}=\{E_{\mathrm{svg},1},\ldots,E_{\mathrm{svg},S}\}\in\mathbb{R}^{S\times d}$ as keys and values. Here, the SVG annotation is treated as structured textual context that describes atom labels, bond relations, and molecular layout, and its representation is embedded in the same semantic space as the language-model input. The attention weight of each vision token over the SVG token sequence is computed as
\begin{equation}
    \beta_{is}
    =
    \mathrm{softmax}_s
    \left(
    \frac{
    \langle W'_q v^{\mathrm{MTM}}_i, W'_k E_{\mathrm{svg},s}\rangle
    }{\sqrt{d_h}}
    \right),
\end{equation}
\begin{equation}
    \tilde{v}_i
    =
    \sum_{s=1}^{S}
    \beta_{is} W'_v E_{\mathrm{svg},s}.
    \label{eq:mgm-attn}
\end{equation}
Through this vision-to-SVG attention, topology-enhanced visual tokens selectively retrieve chemically relevant symbolic cues from the SVG context. The retrieved SVG context $\tilde{v}_i$ is then fused with $v^{\mathrm{MTM}}_i$ through a residual block with layer normalization and feed-forward transformation, producing the updated vision token $v'_i$. The resulting representation $V'$ integrates molecular graph topology from MTM and symbolic chemical semantics from the SVG, and serves as the output of the graph topology adaptation layer.

\subsection{Model Architecture and Pipeline}

Given a molecular image $I \in \mathbb{R}^{H \times W \times 3}$, its image-derived SVG representation $T_{\mathrm{svg}}$, and a natural-language instruction $Q$, MolSight generates a task-dependent response $Y$. Here, $T_{\mathrm{svg}}$ is obtained by vectorizing the same molecular image and contains only visible drawing elements, including atom symbols, bond primitives, and their 2D positions, rather than target SMILES strings, molecular properties, or database-level annotations. Unlike SMILES-based molecular LLMs, MolSight uses these image-side structural cues to align visual representations and capture molecular topology.
ruhe
\textbf{SVG formatting.}
The raw SVG is formatted to retain molecularly relevant structural information while removing contents unrelated to molecular semantics, such as redundant style attributes, rendering metadata, and non-structural elements. This formatting preserves atom labels, bond connectivity, and two-dimensional coordinates in compact form, but substantially shortens the SVG text sequence, thereby reducing the computational burden of SVG-token encoding and vision-to-SVG attention.

MolSight builds on Qwen3-VL as its base VLM. The vision encoder $\mathcal{E}_V$ encodes the molecular image into vision tokens, while the SVG text is processed by the base model's tokenizer and word-embedding layer $\mathcal{E}_T$:
\begin{equation}
    V = \mathcal{E}_V(I) \in \mathbb{R}^{K \times d}, 
    \qquad
    E_{\mathrm{svg}} = \mathcal{E}_T(T_{\mathrm{svg}}) \in \mathbb{R}^{S \times d}.
    \label{eq:encoding}
\end{equation}
The vision tokens are sequentially processed by MTM and MGM:
\begin{equation}
    V_{\mathrm{MTM}} = \Phi_{\mathrm{MTM}}(V),
    \qquad
    V' = \Phi_{\mathrm{MGM}}(V_{\mathrm{MTM}}, E_{\mathrm{svg}}).
    \label{eq:pipeline}
\end{equation}
Finally, the aligned vision tokens $V'$ are concatenated with the instruction tokens $\mathcal{E}_T(Q)$ and passed to the LLM decoder for autoregressive generation. The two newly introduced modules, $\Phi_{\mathrm{MTM}}$ and $\Phi_{\mathrm{MGM}}$, together form the \emph{graph topology adaptation layer} of MolSight, which bridges the vision encoder and the LLM decoder.

\subsection{Training Strategy}

The training objective of MolSight is to enable the model to generate task-relevant responses while learning molecular topology from vision tokens. We use two supervision signals: the standard autoregressive task loss and an edge supervision loss for the MTM edge predictor. The task loss is
\begin{equation}
\mathcal{L}_{\mathrm{task}}
=
-\sum_{t=1}^{|Y|}
\log
P_{\theta}
\left(
y_t
\mid
y_{<t}, V', \mathcal{E}_T(Q)
\right).
\label{eq:task-loss}
\end{equation}
The edge supervision loss constrains the predicted edge logits $Z$ to match the ground-truth bond adjacency matrix $A^*$:
\begin{equation}
\mathcal{L}_{\mathrm{edge}}
=
\mathrm{BCEWithLogits}(Z, A^*).
\label{eq:edge-loss}
\end{equation}
The two losses are used selectively across different training stages.

We adopt a two-stage training strategy, where Stage 1 progressively pretrains the topology adaptation layer through two substages and Stage 2 adapts the model to downstream tasks.

In \textbf{Substage 1}, we activate the MTM and pretrain it on molecular image--task label pairs. The optimization objective is $\mathcal{L}_{\mathrm{task}}+\lambda\mathcal{L}_{\mathrm{edge}}$, where edge supervision enables the MTM edge predictor to learn chemical-bond connectivity from molecular images. 
In \textbf{Substage 2}, we activate the MGM and jointly train it with the MTM to align SVG symbolic semantics with visual topological features, using only $\mathcal{L}_{\mathrm{task}}$.
In \textbf{Stage 2}, we perform downstream task fine-tuning. We freeze the vision encoder and the graph topology adaptation layer, and fine-tune the LLM decoder using LoRA so that the model can adapt to the output format of each downstream task. Detailed pseudocode for the training and inference procedures is provided in Algorithms 1 and 2 in the Appendix~\hyperref[app:algorithms]{A}.
\section{Experiments}\label{experiments}
In this section, we evaluate MolSight from complementary aspects of chemical visual reasoning. We first introduce the experimental setup, followed by performances on four molecular understanding tasks. Finally, we conduct ablation studies to analyze the contribution of each component.

For Stage 1 pretraining, we follow PubChemSTM~\citep{moleculestm} to construct a dataset of 249K molecular image--task label pairs. We remove samples that overlap with downstream task datasets to avoid data leakage. The pretraining tasks include molecular text completion, SMILES reconstruction, and structural feature prediction, providing natural-language, molecular-language, and numerical-feature objectives to train the graph topology adaptation layer.

For downstream evaluation, we compare MolSight with traditional or domain-specific methods, generalist and molecular specialist LLMs/VLMs in all tasks. Task definitions, baselines, and evaluation metrics are introduced in the corresponding subsections, with data processing and training task design detailed in Appendix~\hyperref[app:data_preparation]{B}.
\vspace{-2pt}
\begin{table}[!ht]
\centering
\caption{Performance comparison on image-to-SMILES translation, with baseline results from~\citep{tan2025chemmllm}. Avg Sim denotes Tanimoto similarity. Task-specific OCSR tools are reported as references and are not considered when marking best and second-best results. Best results are in bold and second-best results are underlined.}
\label{tab:img2smiles}
\small
\setlength{\tabcolsep}{6pt}
\resizebox{\columnwidth}{!}{%
\begin{tabular}{lccc}
\toprule
Model 
& Avg Sim $\uparrow$ 
& ACC $\uparrow$ 
& Valid\% $\uparrow$ \\
\midrule

\rowcolor{gray!20}
\multicolumn{4}{l}{\textit{Task-specific OCSR tools}} \\
\hline
MolScribe 
& 0.98 $\pm$ 0.002 
& 0.66 $\pm$ 0.01 
& 96.9\% \\
Decimer 
& 0.97 $\pm$ 0.002 
& 0.78 $\pm$ 0.01 
& 99.9\% \\

\midrule
\rowcolor{gray!20}
\multicolumn{4}{l}{\textit{Generalist VLMs}} \\
\hline
Qwen-VL-7B 
& 0.08 $\pm$ 0.006 
& 0.0 $\pm$ 0.0 
& 8.2\% \\
InternVL-20B 
& 0.09 $\pm$ 0.003 
& 0.0 $\pm$ 0.0 
& 20.7\% \\
LLaVA-7B 
& 0.05 $\pm$ 0.004 
& 0.0 $\pm$ 0.0 
& 11.1\% \\
GPT-4o 
& 0.29 $\pm$ 0.005 
& 0.01 $\pm$ 0.004 
& 74.5\% \\

\midrule
\rowcolor{gray!20}
\multicolumn{4}{l}{\textit{Molecular specialist VLMs}} \\
\hline
ChemVLM-8B 
& 0.55 $\pm$ 0.009 
& 0.11 $\pm$ 0.01 
& 85.2\% \\
ChemMLLM-7B 
& 0.75 $\pm$ 0.009 
& 0.39 $\pm$ 0.01 
& 97.1\% \\
ChemMLLM-34B 
& \underline{0.87 $\pm$ 0.007} 
& \underline{0.56 $\pm$ 0.01} 
& \underline{97.2\%} \\
\rowcolor{blue!8}
\textbf{MolSight-8B} 
& \textbf{0.998 $\pm$ 0.0001} 
& \textbf{0.81 $\pm$ 0.0058} 
& \textbf{99.7\%$\pm 0.04\%$} \\
\rowcolor{blue!8}
\textbf{MolSight-32B} 
& \textbf{0.999 $\pm$ 0.0001} 
& \textbf{0.83 $\pm$ 0.0041} 
& \textbf{99.7\%$\pm 0.04\%$} \\
\bottomrule
\end{tabular}%
}
\end{table}
\begin{table*}[t]
\centering
\caption{ACC Performance comparison on MoleculeQA benchmark, with baseline results from~\citep{moleculeqa}. Best results are in bold and second-best results are underlined.}
\label{tab:moleculeqa}
\setlength{\tabcolsep}{6pt}
\begin{tabular}{lcccccc}
\toprule
Model 
& \# Trainable Params &
Structure (3113) & 
Source (1343) & 
Property  (731) & 
Application (599) & 
Total (5786) \\
\midrule
Random 
& - 
& 24.41 
& 22.30 
& 23.04 
& 24.57 
& 24.03 \\
\hline
\rowcolor{gray!20}
\multicolumn{7}{l}{\textit{Generalist LLMs}} \\
\hline
T5-base 
& 220M 
& 60.42 
& 66.42 
& 45.83 
& \underline{43.74} 
& 58.24 \\
OPT-350M 
& 331M 
& 44.39 
& 60.83 
& 46.24 
& 40.57 
& 48.05 \\
GALACTICA-6.7B 
& 12.5M 
& 32.35 
& 41.92 
& 31.05 
& 28.21 
& 33.96 \\
BLOOM-7.1B 
& 27.5M 
& 35.01 
& 47.51 
& 31.46 
& 33.56 
& 37.31 \\
Pythia-6.9B 
& 29.4M 
& 42.79 
& 58.90 
& 38.58 
& 39.07 
& 45.61 \\
Mol-Instruction-7B 
& 40M 
& 37.46 
& 47.36 
& 32.69 
& 29.88 
& 38.37 \\
Llama-2-13B-chat 
& 63M 
& 34.37 
& 43.86 
& 31.05 
& 29.72 
& 35.67 \\
Vicuna-v1.5-13B 
& 63M 
& 37.01 
& 43.19 
& 30.64 
& 31.55 
& 37.07 \\
Mixtral-8×7B-Instruct-v0.1
& 10-shot 
& 23.32 
& 31.87 
& 32.89 
& 29.96
& 27.79 \\
GPT-4-1106-preview
& 10-shot 
& 60.94 
& 50.19 
& 35.57 
& 43.91
& 53.47 \\
\midrule
\rowcolor{gray!20}
\multicolumn{7}{l}{\textit{Molecular Specialist LLMs}} \\
\hline
MolT5-base 
& 250M 
& 58.01 
& 65.85 
& 45.14 
& 42.24 
& 55.39 \\
MoMu-base 
& 252M 
& 61.58 
& 65.30 
& 43.78 
& 43.07 
& 57.43 \\
BioT5-base 
& 252M 
& 65.98 
& 69.24 
& \underline{49.11} 
& 40.73 
& 62.03 \\
MolCA-125M 
& 100M 
& 65.54 
& 67.34 
& 45.77 
& 40.33 
& 60.30 \\
MolCA-1.3B 
& 110M 
& \underline{71.12} 
& \underline{70.98} 
& 47.81 
& 43.17 
& \underline{64.79} \\
BioMedGPT-LM-7B 
& 40M 
& 54.19 
& 60.01 
& 38.85 
& 40.90 
& 52.23 \\
\rowcolor{blue!8}
\textbf{MolSight-8B} 
& 90.4M 
& \textbf{78.38} 
& \textbf{73.42} 
& \textbf{51.16} 
& \textbf{50.42} 
& \textbf{70.90} \\
\rowcolor{blue!8}
\textbf{MolSight-32B} 
& 272.4M 
& \textbf{80.73} 
& \textbf{76.69} 
& \textbf{54.86} 
& \textbf{51.42} 
& \textbf{73.49} \\
\bottomrule
\end{tabular}
\end{table*}

\subsection{SMILES Translation}\label{sec:smiles_translation}
SMILES translation requires the model to generate the canonical SMILES string of a molecule directly from its image. This task provides the most direct evaluation of MolSight's visual perception: producing a syntactically valid SMILES that matches the ground truth exactly requires the model not only to identify every atom symbol, but also to correctly capture all chemical-bond connectivity, ring closures, and stereochemistry. During Stage 2 fine-tuning, we selected 95k samples from PubChem~\citep{pubchem} for training and 5k disjoint from the training set for testing. We assess prediction quality from three perspectives: Tanimoto similarity over Morgan fingerprints, exact accuracy and SMILES validity.

As shown in Table~\ref{tab:img2smiles}, MolSight outperforms existing VLM methods in the SMILES translation task. The generalist VLMs performed poorly in all metrics, indicating that the standard visual representation is insufficient to restore the molecular structure from chemical images. VLMs adapted for the field of chemistry have improved compared to generalist models, but they still clearly lag behind task-specific OCSR tools~\citep{molscribe, decimer}. This indicates that simple domain adaptation cannot fully solve the problem of molecular structure perception. In contrast, MolSight fills this gap: it not only surpasses all generalist and molecular VLMs, but also exceeds the performance of specialized OCSR tools, while still remaining a unified vision-language model. These results indicate that explicitly adapting visual representations to molecular topologies is crucial for reliably perceiving structures from chemical images. 

Since molecular images may also contain stereochemical cues, we further report additional ACC and Tanimoto metrics that account for stereochemical information in Appendix~\hyperref[app:stereo_analysis]{E.1}.

\subsection{Molecular Captioning}\label{sec:molecular_captioning}
Caption generation requires the model to produce a natural-language content describing the structure, properties, and functions of a molecule from its image. This task evaluates whether MolSight can translate visually perceived molecular structure into chemical semantics. However, regarding the evaluation strategy for this task, our view is consistent with that of MoleculeQA~\citep{moleculeqa} and RTMol~\citep{rtmol}. That is, it is currently popular to directly apply NLP-based metrics such as BLEU and ROUGE, which are \textbf{not sensitive to describing key information}, such as molecular identity. A single numerical difference points to a completely wrong caption, but it has almost no impact on the current metrics. Therefore, for the molecular captioning task, we chose to conduct the test on the MoleculeQA benchmark. This benchmark constructs choice questions distinguished by four types of key information to test whether LLMs truly understand the structure, property, synthesis of molecules.

We evaluate MolSight on the MoleculeQA dataset, which consists of 5,786 samples across four categories: Structure, Source, Property, and Application. The model is trained on a 50k molecular captioning training set, with all LLM-based methods fine-tuned on the same dataset. As shown in Table~\ref{tab:moleculeqa}, MolSight outperforms all generalist and molecular LLMs, demonstrating significant improvements across all four categories. Specifically, MolSight achieves the highest scores in Structure, Source, Property, and Application, surpassing both generalist LLMs and molecular LLMs. Notably, MolSight's performance is achieved with a similar number of trainable parameters as the other LLM-based methods, demonstrating the effectiveness of its topology-aware approach.

To further compare with commonly used molecular captioning protocols, we also evaluate MolSight on the traditional CHEBI-20 benchmark~\citep{chebi20} using NLP-based metrics. As shown in Table~\ref{tab:captioning_nlp_metrics}, MolSight achieves the best performance across BLEU, ROUGE, and METEOR, outperforming both generalist VLMs and molecular specialist VLMs. These results further confirm MolSight's advantage in molecular caption generation.
\begin{table*}[!ht]
\centering
\caption{Additional performance comparison on molecular captioning using NLP-based metrics. Higher values indicate better performance. Best results are in bold and second-best results are underlined.}
\label{tab:captioning_nlp_metrics}
\small
\begin{tabular}{lcccccc}
\toprule
Model 
& BLEU-2$\uparrow$ 
& BLEU-4$\uparrow$ 
& ROUGE-1$\uparrow$ 
& ROUGE-2$\uparrow$ 
& ROUGE-L$\uparrow$ 
& METEOR$\uparrow$ \\
\midrule
\rowcolor{gray!20}
\multicolumn{7}{l}{\textit{Generalist VLMs}} \\
\hline
Qwen-VL-7B 
& 0.09 & 0.01 & 0.32 & 0.07 & 0.22 & 0.19 \\
InternVL-v1.5-20B 
& 0.04 & 0.001 & 0.38 & 0.07 & 0.26 & 0.19 \\
LLaVA-v1.5-7B 
& 0.08 & 0.004 & 0.33 & 0.08 & 0.23 & 0.20 \\
GPT-4o 
& 0.16 & 0.07 & 0.28 & 0.13 & 0.23 & 0.22 \\
\midrule
\rowcolor{gray!20}
\multicolumn{7}{l}{\textit{Molecular specialist VLMs}} \\
\hline
ChemVLM-8B 
& 0.15 & 0.08 & 0.28 & 0.13 & 0.23 & 0.23 \\
ChemMLLM-7B 
& 0.33 & 0.21 & 0.50 & 0.31 & 0.43 & 0.43 \\
ChemMLLM-34B 
& \underline{0.36} 
& \underline{0.25} 
& \underline{0.51} 
& \underline{0.33} 
& \underline{0.45} 
& \underline{0.45} \\
\rowcolor{blue!8}
\textbf{MolSight-8B} 
& \textbf{0.46} 
& \textbf{0.34} 
& \textbf{0.56} 
& \textbf{0.39} 
& \textbf{0.49} 
& \textbf{0.46} \\
\rowcolor{blue!8}
\textbf{MolSight-32B} 
& \textbf{0.48} 
& \textbf{0.36} 
& \textbf{0.61} 
& \textbf{0.44} 
& \textbf{0.54} 
& \textbf{0.52} \\
\bottomrule
\end{tabular}
\end{table*}

\begin{table*}[t]
\centering
\caption{Performance comparison on physicochemical descriptor prediction, with baseline results from~\citep{tan2025chemmllm}. Pearson correlation (higher is better) and MAE (lower is better) are reported for seven molecular descriptors. Task-specific traditional models are reported as references and are not considered when marking best and second-best results.}
\label{tab:descriptor_prediction}
\setlength{\tabcolsep}{3pt}
\resizebox{\textwidth}{!}{
\begin{tabular}{lcccccccccccccccc}
\toprule
\multirow{2}{*}{Method} 
& \multicolumn{2}{c}{MW} 
& \multicolumn{2}{c}{LogP} 
& \multicolumn{2}{c}{TPSA} 
& \multicolumn{2}{c}{HBD} 
& \multicolumn{2}{c}{HBA} 
& \multicolumn{2}{c}{RB} 
& \multicolumn{2}{c}{QED} 
& \multicolumn{2}{c}{Avg} \\
\cmidrule(lr){2-3}
\cmidrule(lr){4-5}
\cmidrule(lr){6-7}
\cmidrule(lr){8-9}
\cmidrule(lr){10-11}
\cmidrule(lr){12-13}
\cmidrule(lr){14-15}
\cmidrule(lr){16-17}
& Pearson$\uparrow$ & MAE$\downarrow$
& Pearson$\uparrow$ & MAE$\downarrow$
& Pearson$\uparrow$ & MAE$\downarrow$
& Pearson$\uparrow$ & MAE$\downarrow$
& Pearson$\uparrow$ & MAE$\downarrow$
& Pearson$\uparrow$ & MAE$\downarrow$
& Pearson$\uparrow$ & MAE$\downarrow$
& Pearson$\uparrow$ & MAE$\downarrow$ \\
\midrule

\rowcolor{gray!20}
\multicolumn{17}{l}{\textit{Task-specific traditional models}} \\
\hline
Chemception 
& 0.87 & 44.75 
& 0.62 & 1.23 
& 0.84 & 14.45 
& 0.81 & 0.58 
& 0.84 & 0.29 
& 0.73 & 7.57 
& -0.004 & 0.40 
& 0.673 & 9.90 \\
Chemprop 
& 0.99 & 1.63 
& 0.98 & 0.16 
& 0.99 & 0.91 
& 0.99 & 0.03 
& 0.99 & 0.03 
& 0.99 & 0.05 
& 0.89 & 0.07 
& 0.974 & 0.41 \\
\midrule

\rowcolor{gray!20}
\multicolumn{17}{l}{\textit{Generalist VLMs}} \\
\hline
Qwen-VL-7B 
& 0.78 & 58.07 
& 0.14 & 1.48 
& 0.19 & 82.83 
& -0.02 & 1.42 
& 0.05 & 30.56 
& 0.19 & 145.40 
& 0.00 & -- 
& 0.190 & -- \\
InternVL-20B 
& 0.59 & 83.60 
& 0.04 & 2.37 
& 0.29 & 28.73 
& 0.03 & 2.24 
& 0.04 & 10.24 
& 0.04 & 43.67 
& 0.003 & 4.01 
& 0.147 & 24.98 \\
LLaVA-7B 
& 0.36 & 115.70 
& -0.003 & 1.61 
& 0.01 & 99.14 
& 0.004 & 3.74 
& 0.04 & 23.71 
& 0.04 & 39.99 
& -0.11 & 24.53 
& 0.049 & 44.06 \\
\midrule

\rowcolor{gray!20}
\multicolumn{17}{l}{\textit{Molecular specialist VLMs}} \\
\hline
ChemVLM-8B 
& 0.84 & 56.94 
& 0.38 & 1.68 
& 0.26 & 53.66 
& 0.49 & 1.35 
& 0.32 & 27.27 
& 0.10 & 45.88 
& -0.003 & 0.24 
& 0.341 & 26.72 \\
ChemMLLM-7B 
& \underline{0.97} & 16.17 
& 0.92 & 0.70 
& 0.97 & 6.06 
& \underline{0.96} & 0.13 
& 0.94 & 0.79 
& 0.94 & 1.62 
& 0.91 & 0.06 
& 0.944 & 3.64 \\
ChemMLLM-34B 
& \underline{0.98} & \underline{11.57} 
& \underline{0.93} & \underline{0.53} 
& \underline{0.98} & \underline{3.54} 
& \underline{0.96} & \underline{0.11} 
& \underline{0.96} & \underline{0.53} 
& \underline{0.97} & \underline{0.75} 
& \underline{0.93} & \underline{0.05} 
& \underline{0.959} & \underline{2.43} \\
\rowcolor{blue!8}
\textbf{MolSight-8B} 
& \textbf{0.996} & \textbf{1.78} 
& \textbf{0.97} & \textbf{0.30} 
& \textbf{0.999} & \textbf{0.96} 
& \textbf{0.996} & \textbf{0.01} 
& \textbf{0.99} & \textbf{0.09} 
& \textbf{0.99} & \textbf{0.25} 
& \textbf{0.94} & \textbf{0.05} 
& \textbf{0.984} & \textbf{0.49} \\
\rowcolor{blue!8}
\textbf{MolSight-32B} 
& \textbf{0.998} & \textbf{1.57} 
& \textbf{0.96} & \textbf{0.32} 
& \textbf{0.998} & \textbf{1.29} 
& \textbf{0.996} & \textbf{0.02} 
& \textbf{0.99} & \textbf{0.09} 
& \textbf{0.99} & \textbf{0.24} 
& \textbf{0.95} & \textbf{0.04} 
& \textbf{0.983} & \textbf{0.51} \\
\bottomrule
\end{tabular}
}
\end{table*}
\subsection{Descriptor Estimation}\label{sec:descriptor_prediction}
Descriptor estimation refers to enabling the model to predict the physicochemical properties of the molecule. We selected seven properties related to structure and function: molecular weight, Octanol-water partition, topological polar surface area, hydrogen bond donor/acceptor, rotatable bonds, and drug-likeness, The definition of descriptors is in Appendix~\hyperref[app:tasks_metrics]{C}. Consistent with other baselines, we trained on a 95k dataset from PubChem~\citep{pubchem} and tested on 5k samples, and report the mean absolute error (MAE) and Pearson correlation between predicted values and ground truth.

As shown in Table~\ref{tab:descriptor_prediction}, MolSight performs excellently across all physicochemical descriptor prediction tasks. Compared to all generalist and molecular specialist VLMs, MolSight's performance significantly leads. For properties like MW, TPSA, and HBD, the Pearson correlation is nearly 1, suggesting an almost perfect positive correlation. Moreover, MolSight achieves competitive results with task-specific methods such as Chemprop. Remarkably, these results are achieved with a lightweight fine-tune after topology adaptation.

\begin{table*}[t]
\centering
\caption{Bioactivity prediction performance comparison on MolVision, with baseline results from~\citep{molvision}. All methods are fine-tuned. Classification tasks report ACC and F1 in the format ACC(F1), where higher is better. ESOL-V reports RMSE, where lower is better. Best results are in bold and second-best results are underlined.}
\label{tab:bioactivity_prediction}
\setlength{\tabcolsep}{5pt}
\resizebox{\textwidth}{!}{
\begin{tabular}{lccccccc}
\toprule
Model 
& BACE-V$\uparrow$ 
& BBBP-V$\uparrow$ 
& HIV-V$\uparrow$ 
& ClinTox-V$\uparrow$ 
& Tox21-V$\uparrow$ 
& Average$\uparrow$ 
& ESOL-V(RMSE)$\downarrow$ \\
\midrule

\rowcolor{gray!20}
\multicolumn{8}{l}{\textit{Traditional methods}} \\
\hline
RF 
& 0.79(0.76) 
& 0.82(0.88) 
& 0.87(0.52) 
& 0.85(0.46) 
& 0.83(0.26) 
& 0.83(0.57) 
& -- \\
XGBoost 
& 0.81(0.77) 
& 0.85(0.90) 
& 0.87(0.55) 
& 0.88(0.62) 
& 0.84(0.33) 
& 0.85(0.63) 
& -- \\

\midrule
\rowcolor{gray!20}
\multicolumn{8}{l}{\textit{SMILES-based LLMs}} \\
\hline
ChemLLM 
& 0.18(0.12) 
& 0.12(0.08) 
& 0.19(0.09) 
& 0.21(0.13) 
& 0.18(0.09) 
& 0.17(0.10) 
& -- \\
MolCA 
& 0.79(0.73) 
& 0.74(0.72) 
& 0.89(0.84) 
& \textbf{0.93}(0.84) 
& 0.80(0.72) 
& 0.83(0.77) 
& -- \\

\midrule
\rowcolor{gray!20}
\multicolumn{8}{l}{\textit{Image-based VLMs}} \\
\hline
BLIP-2 
& \underline{0.86}(\underline{0.83}) 
& \textbf{0.93}(\textbf{0.96}) 
& \underline{0.92}(0.76) 
& \underline{0.89}(\textbf{0.93}) 
& \textbf{0.99}(0.80) 
& \underline{0.92}(0.86) 
& 1.764 \\
LLaVA-1.5-13B 
& 0.84(0.83) 
& 0.86(0.88) 
& 0.80(0.81) 
& 0.70(0.72) 
& 0.92(0.93) 
& 0.82(0.83) 
& 2.229 \\
LLaMA-Adapter-v2-7B 
& 0.52(0.48) 
& 0.45(0.46) 
& 0.43(0.42) 
& 0.58(0.62) 
& 0.68(0.69) 
& 0.53(0.53) 
& 4.032 \\
CogVLM 
& 0.72(0.71) 
& 0.78(0.82) 
& 0.85(0.83) 
& 0.88(0.90) 
& 0.93(0.93) 
& 0.83(0.84) 
& \underline{1.102} \\
Qwen-VL 
& 0.78(0.78) 
& 0.70(0.72) 
& 0.60(0.61) 
& 0.71(0.64) 
& 0.75(0.64) 
& 0.71(0.68) 
& 2.192 \\
mPlugOWL2 
& 0.86(0.82) 
& 0.90(0.88) 
& 0.90(\underline{0.91}) 
& 0.89(0.92) 
& 0.94(\textbf{0.96}) 
& 0.89(\underline{0.89}) 
& 1.291 \\
\rowcolor{blue!8}
\textbf{MolSight-8B} 
& \textbf{0.90}(\textbf{0.90}) 
& \textbf{0.93}(\textbf{0.96}) 
& \textbf{0.96}(\textbf{0.94}) 
& 0.88(\textbf{0.93}) 
& \textbf{0.99}(\underline{0.84}) 
& \textbf{0.93}(\textbf{0.91}) 
& \textbf{0.899} \\
\bottomrule
\end{tabular}
}
\end{table*}
\subsection{Bioactivity Prediction}\label{sec:bioactivity_prediction}
Finally, we evaluate our MolSight's performance on bioactivity property prediction tasks. We chose the widely used benchmark, MoleculeNet~\citep{moleculenet}, which contains bioactivity tasks that evaluate a model's ability to predict the biological activity of molecules across various target proteins and compounds. MolVision~\citep{molvision} integrates the performance of common VLMs on various MoleculeNet tasks. We conducted a fair and consistent comparison on MolVision, using the same training set and LoRA fine-tuning strategy, and reported the comparison results on both classification and regression tasks. For classification tasks, we computed ACC and F1 score, while for regression tasks, we calculated RMSE.

The six bioactivity tasks above include BACE predicting the inhibitory activity of molecules against $\beta$-secretase, BBBP predicting the ability of molecules to cross the blood-brain barrier, HIV predicting the inhibitory activity of molecules against the human immunodeficiency virus, ClinTox predicting the clinical toxicity of molecules, Tox21 evaluating the toxicity and environmental risks of molecules, and ESOL predicting the solubility of molecules. For complete definitions, see the Appendix~\hyperref[app:tasks_metrics]{C}. As shown in Table~\ref{tab:bioactivity_prediction}, MolSight performs excellently in the MolVision bioactivity prediction tasks, surpassing all other methods. Compared to other fine-tuned SMILES-based molecular LLMs and generalist VLMs, MolSight achieves a significant lead in overall performance and achieves the best results in 5 out of 6 tasks.

\subsection{Ablation Studies}
In this section, we conduct ablation studies to systematically analyze the factors that contribute to MolSight's performance.
\begin{table}[!ht]
\centering
\caption{Ablation results on SMILES translation. Avg. Sim denotes Tanimoto similarity. Higher is better for all metrics.}
\label{tab:ablation_img2smiles}
\small
\setlength{\tabcolsep}{8pt}
\begin{tabular}{lccc}
\toprule
Model & Avg. Sim $\uparrow$& ACC $\uparrow$& Valid (\%) $\uparrow$\\
\midrule
\rowcolor{blue!8}
\textbf{MolSight-8B} 
& \textbf{0.998} 
& \textbf{0.807}
& 99.7\% \\
w/o MTM 
& 0.989 
& 0.776 
& 99.6\% \\
w/o MGM 
& 0.993 
& 0.788 
& \textbf{99.8\%} \\
\makecell[l]{w/o MTM+MGM\\(direct fine-tune)}
& 0.931 
& 0.500 
& 85.8\% \\
w/o 2-substage 
& 0.974 
& 0.751 
& 99.6\% \\
\bottomrule
\end{tabular}
\end{table}

First, we examine the contribution of MolSight’s core modules and training strategy. As shown in Table~\ref{tab:ablation_img2smiles}, the MolSight model achieves the best Avg. Sim and ACC, indicating that the cooperation between MTM and MGM is critical for accurate molecular structure recovery. Removing either MTM or MGM leads to performance degradation: removing MTM weakens the model’s ability to capture bond connectivity and topological relationships, while removing MGM affects the alignment between visual representations and SVG structural semantics. Furthermore, when both MTM and MGM are removed, which is equivalent to directly fine-tuning the base VLM, the task performance drops. This indicates that the gains brought by our visual topology information cannot be achieved simply through downstream fine-tuning. In addition, merging the two substages of Stage 1 and training both modules simultaneously also degrades performance, suggesting that progressively activating the topology adaptation layer helps the model learn topological semantics from molecular images more stably.

\begin{table}[!ht]
\centering
\caption{Ablation study on LoRA fine-tuning hyperparameters. We report representative metrics for SMILES translation, molecular captioning, and descriptor prediction. Higher is better for ACC, while lower is better for Avg. MAE.}
\label{tab:ablation_lora}
\small
\setlength{\tabcolsep}{8pt}
\begin{tabular}{lccc}
\toprule
LoRA Setting 
& \makecell{SMILES\\ACC $\uparrow$} 
& \makecell{Captioning\\Total ACC $\uparrow$} 
& \makecell{Property\\Avg. MAE $\downarrow$} \\
\midrule
\rowcolor{blue!8}
$r=32,\ \alpha=64$ 
& \textbf{0.807} 
& \textbf{0.709} 
& \textbf{0.49} \\
$r=64,\ \alpha=128$ 
& 0.805 
& 0.638 
& 0.94 \\
\bottomrule
\end{tabular}
\end{table}
Second, we evaluate multiple LoRA hyperparameter settings to select an appropriate fine-tuning scale. As shown in Table~\ref{tab:ablation_lora}, compared with the larger LoRA configuration, $r=32,\alpha=64$ achieves more stable overall performance across the three tasks. This suggests that once MolSight has learned structured visual representations through the topology adaptation layer, further increasing the LoRA rank and scaling factor does not necessarily bring additional gains and may instead reduce downstream generalization.

\begin{table}[!ht]
\centering
\caption{Ablation study on base-model scale. We compare MolSight with 8B and 4B backbones on image-to-SMILES translation. Higher is better for all metrics.}
\label{tab:ablation_scale}
\small
\setlength{\tabcolsep}{10pt}
\begin{tabular}{lccc}
\toprule
Model & Avg. Sim & ACC & Valid (\%) \\
\midrule
\rowcolor{blue!8}
\textbf{MolSight-8B} 
& \textbf{0.997} 
& \textbf{0.807} 
& \textbf{99.7\%} \\
MolSight-4B 
& 0.995 
& 0.802 
& 99.2\% \\
\bottomrule
\end{tabular}
\end{table}
Third, motivated by the observed gains from scaling the backbone to 32B and by the LoRA hyperparameter study, we further examine whether MolSight suffers a significant performance collapse when using a smaller VLM backbone. As shown in Table~\ref{tab:ablation_scale}, MolSight-8B still achieves the best results across all metrics, suggesting that sufficient base-model capacity remains important for learning structured visual representations. Nevertheless, MolSight-4B does not exhibit a clear performance collapse, indicating that MolSight can maintain strong performance under reduced backbone capacity, even outperforming the 34B ChemMLLM baseline.

Finally, Table~\ref{tab:ablation_img2smiles} evaluates the contribution of each design choice by removing key components and training stages. However, this setup also removes the training data associated with the corresponding stages, raising the concern that the performance drop may be caused by reduced data scale rather than the module design or training strategy. To control for this factor, we use the Stage 1 training data as additional fine-tuning data for SMILES translation, ensuring that the original model and the ablated model are trained on the same data. As shown in Table~\ref{tab:data_scale_ablation}, simply increasing the data scale cannot achieve the same gains as the full MolSight framework.

\begin{table}[H]
\centering
\caption{Ablation study on the effect of training data scale for SMILES translation.}
\label{tab:data_scale_ablation}
\small
\begin{tabular}{lccc}
\toprule
Setting & Avg. Sim$\uparrow$ & ACC$\uparrow$ & Valid (\%)$\uparrow$ \\
\midrule
\rowcolor{blue!8}
\textbf{MolSight} 
& \textbf{0.998} 
& \textbf{0.807} 
& \textbf{99.7\%} \\
Direct FT w/ stage 1 data
& 0.952
& 0.650
& 98.6\% \\
\bottomrule
\end{tabular}
\end{table}

Additional statistical analyses and ablations on other components, input formats, hyperparameters, and data scales are reported in Appendix~\hyperref[app:statistical_analysis]{E} and ~\hyperref[app:additional_ablation]{G}, respectively.

\section*{Future Work}
A promising direction is to design a dedicated structural tokenizer for SVG inputs. Compared with general text tokenizers, such a tokenizer could better encode atom labels, bond connectivity, and spatial layout in molecular SVGs, thereby improving the effectiveness of vision-to-SVG cross-attention. It may also reduce the dependence on raw SVG syntax and provide more compact and chemically meaningful structure-aware tokens for molecular visual grounding. In addition, although MolSight is designed as a modular topology adaptation framework, its current implementation is still closely tied to the default Qwen3VL backbone. Due to differences in visual token organization, multimodal fusion mechanisms, and vision-language alignment strategies, backbone-specific adaptation may still be needed for different VLM architectures.

\section{Conclusion}\label{conclusion}
In this work, we propose MolSight, a graph-aware vision-language framework for molecular image understanding. To address the difficulty of existing VLMs in capturing chemical structural semantics from molecular images, MolSight introduces a topology adaptation layer that learns molecular graph adjacency from vision tokens and aligns visual representations with structured symbolic annotations in SVG. Experiments show that MolSight achieves substantial improvements across multiple levels of molecular visual understanding, including structure perception, textual description, property prediction, and functional prediction, broadly reaching state-of-the-art performance. Overall, MolSight demonstrates that explicitly modeling molecular topology and injecting it into visual representations is an effective path toward reliable molecular image understanding in VLMs, providing a new practical foundation for a unified chemical image understanding interface in multi-modal LLM-driven drug discovery and molecular design.



\section*{Ethics and Privacy Statement}
This work does not involve human subjects, personal data, or user-sensitive information. The experiments use public molecular datasets and focus on computational molecular understanding tasks. MolSight is intended to improve molecular representation understanding rather than provide autonomous chemical synthesis instructions or safety-critical decisions. Its outputs should be regarded as computational predictions and validated by domain experts before use in real-world chemical experiments.

\bibliographystyle{ACM-Reference-Format}
\bibliography{references}

\appendix

\newpage
\clearpage
\appendix

\section*{Appendix}
\label{appendix}

\phantomsection
\subsection*{A. Algorithms}\label{app:algorithms}\mbox{}\\
Following the content of ~\hyperref[methods]{Methods}, we summarize MolSight's training architecture into the following three algorithms.

\begin{algorithm}[!ht]
\caption{MolSight Stage 1, Substage 1}
\label{alg:training_substage1}
\begin{algorithmic}[1]
\Require Molecular image $I$, SVG text $T_{\mathrm{svg}}$, instruction $Q$, target response $Y$, ground-truth adjacency matrix $A^*$, loss weight $\lambda$
\Ensure Base VLM with initialized MTM and MGM

\State Encode the molecular image and SVG text:
\[
    V = E_V(I), \qquad E_{\mathrm{svg}} = E_T(T_{\mathrm{svg}})
\]
\Comment{$E_V$ and $E_T$ are frozen}

\State For each vision token $v_i$, compute source and target edge representations:
\[
    h_i^{\mathrm{src}} = W_s v_i, 
    \qquad 
    h_i^{\mathrm{dst}} = W_d v_i .
\]

\State Compute symmetric edge logits and soft adjacency:
\[
    Z_{ij}
    =
    \frac{1}{2\sqrt{d_e}}
    \left(
    \langle h_i^{\mathrm{src}}, h_j^{\mathrm{dst}} \rangle
    +
    \langle h_j^{\mathrm{src}}, h_i^{\mathrm{dst}} \rangle
    \right),
    \qquad
    \hat{A}_{ij}=\sigma(Z_{ij}).
\]

\State Use $\hat{A}$ as a soft topological mask in graph attention:
\[
    \alpha_{ij}
    =
    \mathrm{softmax}_{j}
    \left(
    \frac{\langle W_q v_i, W_k v_j\rangle}{\sqrt{d_h}}
    \cdot \hat{A}_{ij}
    -
    C(1-\hat{A}_{ij})
    \right).
\]

\State Compute topology-aware visual tokens:
\[
    v_i^{\mathrm{MTM}}
    =
    \mathrm{GATBlock}
    \left(
    v_i,
    \sum_{j=1}^{K}
    \alpha_{ij} W_v v_j
    \right).
\]

\State Pass $V_{\mathrm{MTM}}$ through the frozen MGM:
\[
    V' = \Phi_{\mathrm{MGM}}(V_{\mathrm{MTM}}, E_{\mathrm{svg}})
\]
\Comment{$\Phi_{\mathrm{MGM}}$ is frozen}

\State Compute the task loss:
\[
    \mathcal{L}_{\mathrm{task}}
    =
    -\sum_{t=1}^{|Y|}
    \log P_{\theta}
    \left(
    y_t \mid y_{<t}, V', E_T(Q)
    \right)
\]
\Comment{$P_\theta$ is frozen}

\State Compute the edge supervision loss:
\[
    \mathcal{L}_{\mathrm{edge}}
    =
    \mathrm{BCEWithLogits}(Z,A^*).
\]

\State Optimize with:
\[
    \mathcal{L}
    =
    \mathcal{L}_{\mathrm{task}}
    +
    \lambda \mathcal{L}_{\mathrm{edge}}.
\]

\State \Return $\mathcal{M}_{\theta_{\mathrm{sub1}}}$

\end{algorithmic}
\end{algorithm}


\begin{algorithm}[!ht]
\caption{MolSight Stage 1, Substage 2}
\label{alg:training_substage2}
\begin{algorithmic}[1]
\Require Molecular image $I$, SVG text $T_{\mathrm{svg}}$, instruction $Q$, target response $Y$
\Ensure $\mathcal{M}_{\theta_{\mathrm{sub1}}}$

\State Encode the molecular image and SVG text:
\[
    V = E_V(I), 
    \qquad 
    E_{\mathrm{svg}} = E_T(T_{\mathrm{svg}})
\]
\Comment{$E_V$ and $E_T$ are frozen}

\State Obtain topology-aware vision tokens with the trained MTM:
\[
    V_{\mathrm{MTM}} = \Phi_{\mathrm{MTM}}(V)
\]
\Comment{MTM is frozen}

\State Compute vision-to-SVG attention weights:
\[
    \beta_{is}
    =
    \mathrm{softmax}_{s}
    \left(
    \frac{
    \langle W'_q v^{\mathrm{MTM}}_i, W'_k E_{\mathrm{svg},s} \rangle
    }{\sqrt{d_h}}
    \right).
\]

\State Aggregate SVG symbolic information for each vision token:
\[
    \tilde{v}_i
    =
    \sum_{s=1}^{S}
    \beta_{is} W'_v E_{\mathrm{svg},s}.
\]

\State Compute grounded visual tokens:
\[
    v'_i
    =
    \mathrm{FFN}
    \left(
    \mathrm{LN}
    \left(
    v^{\mathrm{MTM}}_i
    +
    W'_o \tilde{v}_i
    \right)
    \right),
    \qquad
    V'=\{v'_i\}_{i=1}^{K}.
\]

\State Compute the task loss:
\[
    \mathcal{L}
    =
    \mathcal{L}_{\mathrm{task}}
    =
    -\sum_{t=1}^{|Y|}
    \log P_{\theta}
    \left(
    y_t \mid y_{<t}, V', E_T(Q)
    \right)
\]
\Comment{LLM decoder is frozen}

\State Update the trainable parameters using $\mathcal{L}$.

\State \Return $\mathcal{M}_{\theta_{\mathrm{sub2}}}$

\end{algorithmic}
\end{algorithm}


\begin{algorithm}[!ht]
\caption{MolSight Stage 2 Downstream Fine-tuning}
\label{alg:training_stage2}
\begin{algorithmic}[1]
\Require Downstream molecular image $I$, SVG text $T_{\mathrm{svg}}$, instruction $Q$, target response $Y$
\Ensure $\mathcal{M}_{\theta_{\mathrm{sub2}}}$

\State Encode the molecular image and SVG text:
\[
    V = E_V(I), 
    \qquad 
    E_{\mathrm{svg}} = E_T(T_{\mathrm{svg}})
\]
\Comment{$E_V$ and $E_T$ are frozen}

\State Obtain topology-aware vision tokens with the trained MTM:
\[
    V_{\mathrm{MTM}} = \Phi_{\mathrm{MTM}}(V)
\]
\Comment{MTM is frozen}

\State Align topology-aware vision tokens with SVG symbolic annotations:
\[
    V' = \Phi_{\mathrm{MGM}}(V_{\mathrm{MTM}}, E_{\mathrm{svg}})
\]
\Comment{MGM is frozen}

\State Construct the multimodal input to the LLM decoder:
\[
    X = [V'; E_T(Q)].
\]

\State Generate the task-specific response autoregressively:
\[
    P_{\theta}
    \left(
    y_t \mid y_{<t}, X
    \right)
    =
    P_{\theta}
    \left(
    y_t \mid y_{<t}, V', E_T(Q)
    \right).
\]

\State Compute the downstream task loss:
\[
    \mathcal{L}
    =
    \mathcal{L}_{\mathrm{task}}
    =
    -\sum_{t=1}^{|Y|}
    \log P_{\theta}
    \left(
    y_t \mid y_{<t}, V', E_T(Q)
    \right).
\]

\State Update the LoRA parameters in the LLM decoder using $\mathcal{L}$.

\State \Return $\mathcal{M}_{\theta}$

\end{algorithmic}
\end{algorithm}

\phantomsection
\subsection*{B. Data Preparation}\label{app:data_preparation}\mbox{}\\

During pretraining data processing, we first perform deduplication to avoid data leakage. Specifically, molecular SMILES strings are canonicalized and compared against the molecules in downstream test sets, and any overlapping samples are removed from the training set. We then parse valid SMILES strings with RDKit and convert the molecular structures into two-dimensional molecular graph representations. Based on the canonical SMILES, RDKit renders each molecule into both a PNG image and a vectorized molecular structure representation in SVG format. The sizes of the PNG and SVG outputs are controlled by a unified image size hyperparameter, which is set to $512 \times 512$ by default, ensuring that all molecular visual inputs share a consistent spatial resolution. Below, we describe the design strategies of the three pretraining tasks in detail.

\textbf{SMILES reconstruction task.} The second pretraining task aims to train the model to recover molecular language from partially corrupted SMILES while grounding the prediction in the molecular image and SVG structure. We first tokenize each canonical SMILES using a chemistry-aware tokenizer that preserves bracketed atoms, halogens, chirality markers, ring indices, bonds, branches, and aromatic atoms as meaningful units. We then apply four masking strategies: random token masking, substructure masking, stereochemistry masking, and grammar-symbol masking. Random masking removes a fixed portion of SMILES tokens; substructure masking targets functional groups, rings, or branches; stereochemistry masking focuses on symbols such as @, @@, /, and \textbackslash; and grammar masking hides syntactic elements such as parentheses and ring numbers. These strategies are sampled with weights of 30\%, 30\%, 20\%, and 20\%, respectively, so that the model learns not only local token recovery but also chemically meaningful structure completion. The target output is always the original complete SMILES.

\textbf{Structural feature prediction task.} The third pretraining task directly supervises the model to recognize chemically interpretable structural features from molecular images. For each valid molecule, RDKit is used to identify functional groups through predefined SMARTS patterns, including hydroxyl, carboxyl, amine, amide, ester, ether, aldehyde, ketone, nitro, sulfonyl, phosphate, halides, nitrile, alkene, and alkyne. In addition, ring systems are categorized by ring size, aromaticity, and heteroatom composition, producing labels such as benzene, pyridine, pyrimidine, pyrrole, furan, thiophene, cyclohexane, and other heterocycles. The generated answer also includes the number of chiral centers and heavy atoms. This task provides explicit supervision for functional groups, ring structures, stereochemical centers, and basic molecular size, complementing the description and SMILES reconstruction tasks with structured chemical feature labels.

\textbf{Molecular description task.} The first pretraining task is designed to strengthen MolSight's ability to connect molecular visual structures with natural-language chemical semantics. For molecules with available textual descriptions, we construct two types of instruction-response pairs. In the generation mode, the model is asked to generate a complete molecular description from the molecular image and its SVG annotation. In the completion mode, we mask key information in the original description and ask the model to recover the complete description based on the visual and SVG inputs. The masked spans cover chemically meaningful content such as compound names, functional groups, physicochemical property words, molecular relationships, and numerical information. The two modes are sampled with a default ratio of 60\% generation and 40\% completion, encouraging the model to learn both global molecule-level descriptions and fine-grained semantic recovery from structural evidence.

\textbf{Data resources and sizes.}
Table~\ref{tab:dataset_summary} summarizes the dataset resources and sizes used for training at each stage and for evaluation on downstream tasks.
\begin{table*}[!ht]
\centering
\caption{Dataset statistics for different stages and tasks. }
\label{tab:dataset_summary}
\begin{tabular}{lcccc}
\toprule
& Dataset & Total & Training & Test \\
\midrule
Training Stage 1 & PubChemSTM~\citep{moleculestm} & 249,545 & 249,545 & - \\
SMILES Translation & PubChem~\citep{pubchem} & 100,000 & 95,000 & 5,000 \\
Molecular Captioning & MoleculeQA~\citep{moleculeqa} & 55,779 & 49,993 & 5,786 \\
Descriptor Estimation & PubChem~\citep{pubchem} & 100,000 & 95,000 & 5,000 \\
Bioactivity Prediction & MolVision~\citep{molvision} & 55,140 & 44,112 & 11,028 \\
Additional Molecular captioning & \makecell{CHEBI-20~\citep{chebi20} \\ Mol-Instructions~\citep{mol-instructions}} & 73,099 & 69,799 & 3300 \\
\bottomrule
\end{tabular}
\end{table*}

\phantomsection
\subsection*{C. Tasks and Metrics}\label{app:tasks_metrics}\mbox{}\\

\phantomsection
\subsubsection*{C.1 Task Definitions}\label{app:task_definitions}\mbox{}\\

This section provides detailed definitions of the physicochemical descriptors and bioactivity tasks used in our property and functional prediction experiments.

\textbf{Physicochemical descriptors.}
\begin{itemize}
    \item \textbf{MW} denotes molecular weight, defined as the sum of the atomic weights of all atoms in a molecule.

    \item \textbf{LogP} denotes the octanol--water partition coefficient, defined as the logarithm of the ratio of a compound's concentration in octanol to that in water at equilibrium. It measures molecular hydrophobicity or lipophilicity.

    \item \textbf{TPSA} denotes topological polar surface area, defined as the surface area contributed by polar atoms, mainly oxygen, nitrogen, and their attached hydrogens. It is widely used to characterize molecular polarity and is related to hydrogen bonding, permeability, and oral bioavailability.

    \item \textbf{HBD} denotes the number of hydrogen-bond donors in a molecule, typically counting atoms or groups that can donate a hydrogen atom to form a hydrogen bond, such as hydroxyl or amine groups. It is an important descriptor for molecular interaction and drug-likeness.

    \item \textbf{HBA} denotes the number of hydrogen-bond acceptors in a molecule, typically counting electronegative atoms that can accept a hydrogen bond, such as oxygen and nitrogen atoms with available lone pairs. It reflects the molecule's capacity to participate in intermolecular interactions.

    \item \textbf{RB} denotes the number of rotatable bonds, defined as the number of single non-ring bonds around which rotation is allowed.

    \item \textbf{QED} denotes the quantitative estimate of drug-likeness, a desirability-based score that summarizes how similar a molecule's physicochemical properties are to those of known oral drugs. It integrates multiple molecular properties, including molecular weight, LogP, topological polar surface area, hydrogen-bond donors and acceptors, aromatic rings, rotatable bonds, and structural alerts, into a single drug-likeness score.
\end{itemize}

\textbf{Bioactivity tasks.}
\begin{itemize}
    \item \textbf{BACE-V} contains quantitative IC50 values and binary labels for inhibitors of human $\beta$-secretase 1 (BACE-1). In our setting, it is used as a classification task to predict whether a molecule shows inhibitory activity against BACE-1.

    \item \textbf{BBBP-V} provides binary labels indicating whether a molecule can penetrate the blood--brain barrier, evaluating molecular permeability related to central nervous system availability.

    \item \textbf{HIV-V} contains experimentally measured abilities of molecules to inhibit HIV replication. It is used as a classification task to predict whether a molecule exhibits anti-HIV activity.

    \item \textbf{ClinTox-V} contains qualitative labels for FDA-approved drugs and drugs that failed clinical trials due to toxicity, evaluating whether a molecule is associated with clinical toxicity.

    \item \textbf{Tox21-V} provides qualitative toxicity measurements on multiple biological targets, including nuclear receptor signaling and stress response pathways, evaluating molecular toxicity-related biological effects.

    \item \textbf{ESOL-V} contains water solubility data for common organic small molecules, typically reported as log solubility. It is used as a regression task to predict molecular aqueous solubility.
\end{itemize}

\phantomsection
\subsubsection*{C.2 Evaluation Metrics}\label{app:evaluation_metrics}\mbox{}\\

\textbf{Molecular fingerprint similarity.}
For image-to-SMILES translation, we measure the structural similarity between the predicted molecule and the ground-truth molecule using Tanimoto similarity over Morgan fingerprints. Given two molecular fingerprints $f_{\mathrm{pred}}$ and $f_{\mathrm{gt}}$, the Tanimoto similarity is defined as
\begin{equation}
    \mathrm{Tanimoto}(f_{\mathrm{pred}}, f_{\mathrm{gt}})
    =
    \frac{
    |f_{\mathrm{pred}} \cap f_{\mathrm{gt}}|
    }{
    |f_{\mathrm{pred}} \cup f_{\mathrm{gt}}|
    }.
\end{equation}
A higher Tanimoto similarity indicates that the predicted molecule is structurally closer to the ground truth.

\textbf{Validity.}
Validity measures whether the generated SMILES string can be successfully parsed into a chemically valid molecule by RDKit. Given $N$ generated molecules, validity is computed as
\begin{equation}
    \mathrm{Validity}
    =
    \frac{1}{N}
    \sum_{i=1}^{N}
    \mathbb{I}
    \left[
    \mathrm{RDKitParse}(\hat{y}_i) \neq \varnothing
    \right],
\end{equation}
where $\hat{y}_i$ is the generated SMILES string and $\mathbb{I}[\cdot]$ is the indicator function.

\textbf{Classification metrics.}
For bioactivity classification tasks, we report accuracy (ACC) and F1 score. ACC measures the proportion of correctly classified samples, while F1 is the harmonic mean of precision and recall. Higher ACC and F1 indicate better classification performance.

\textbf{Regression metrics.}
For regression tasks, we report mean absolute error (MAE) and root mean squared error (RMSE):
\begin{equation}
    \mathrm{MAE}
    =
    \frac{1}{N}
    \sum_{i=1}^{N}
    |y_i - \hat{y}_i|,
    \quad
    \mathrm{RMSE}
    =
    \sqrt{
    \frac{1}{N}
    \sum_{i=1}^{N}
    (y_i - \hat{y}_i)^2
    }.
\end{equation}
Lower MAE and RMSE indicate better regression performance.

\textbf{Captioning metrics.}
For molecular captioning, we report BLEU-$k$, ROUGE-$k$, and METEOR. BLEU-$k$ measures the modified $n$-gram precision between the generated caption and the reference caption up to $k$-grams. ROUGE-$k$ measures the $k$-gram overlap between the generated caption and the reference caption, with a stronger emphasis on recall. METEOR evaluates caption quality by combining unigram precision, unigram recall, and an alignment-based penalty. Higher BLEU-$k$, ROUGE-$k$, and METEOR indicate better agreement with the reference captions.

\phantomsection
\subsection*{D. Additional Captioning Result}\label{app:additional_captioning}\mbox{}\\

\begin{figure}[!ht]
\centering
\includegraphics[width=0.5\textwidth]{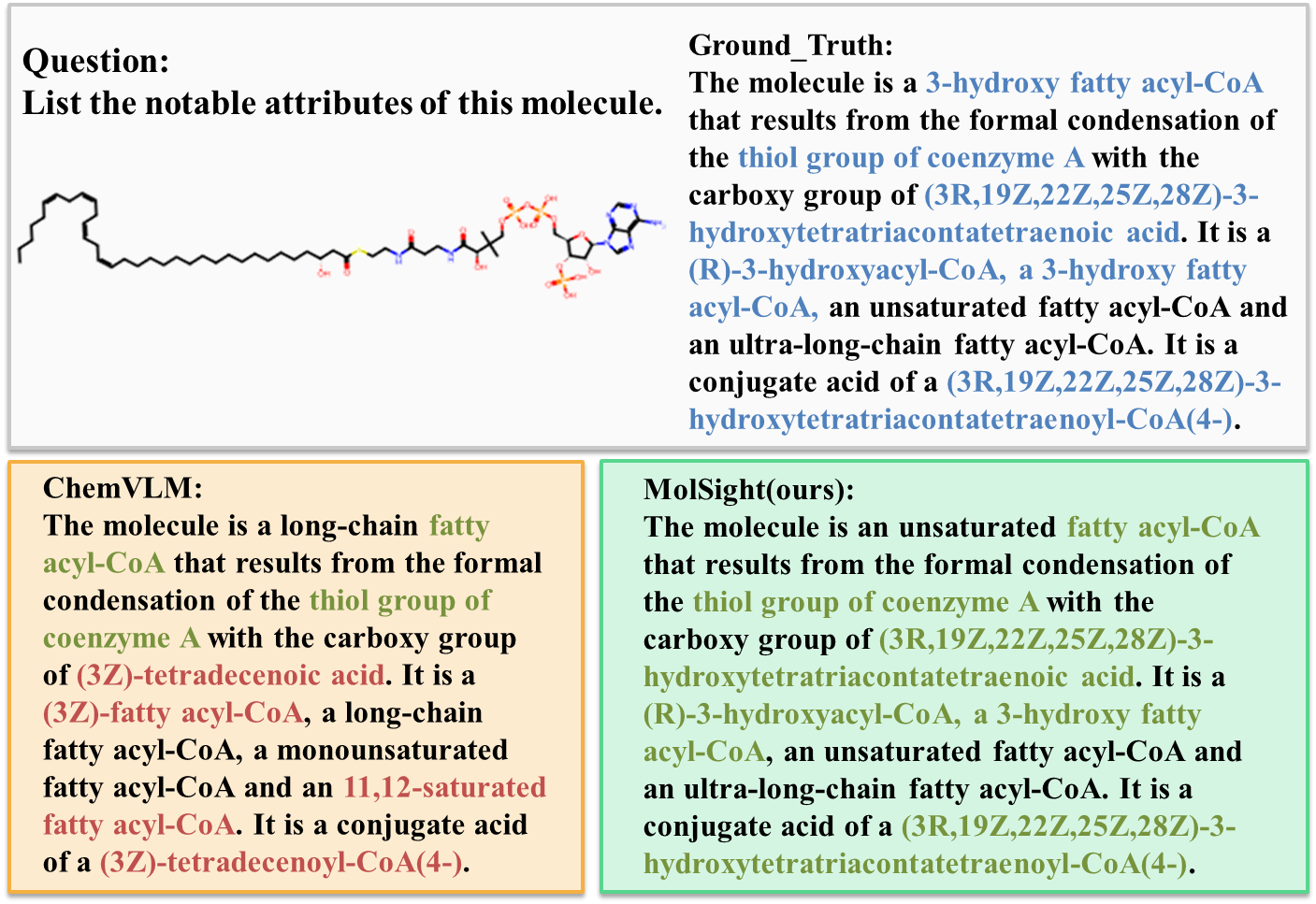}
\caption{Representative molecular captioning comparison between MolSight and ChemVLM.}
\label{fig:caption_case_chemvlm}
\Description{ccc1}
\end{figure}

\begin{figure}[!ht]
\centering
\includegraphics[width=0.5\textwidth]{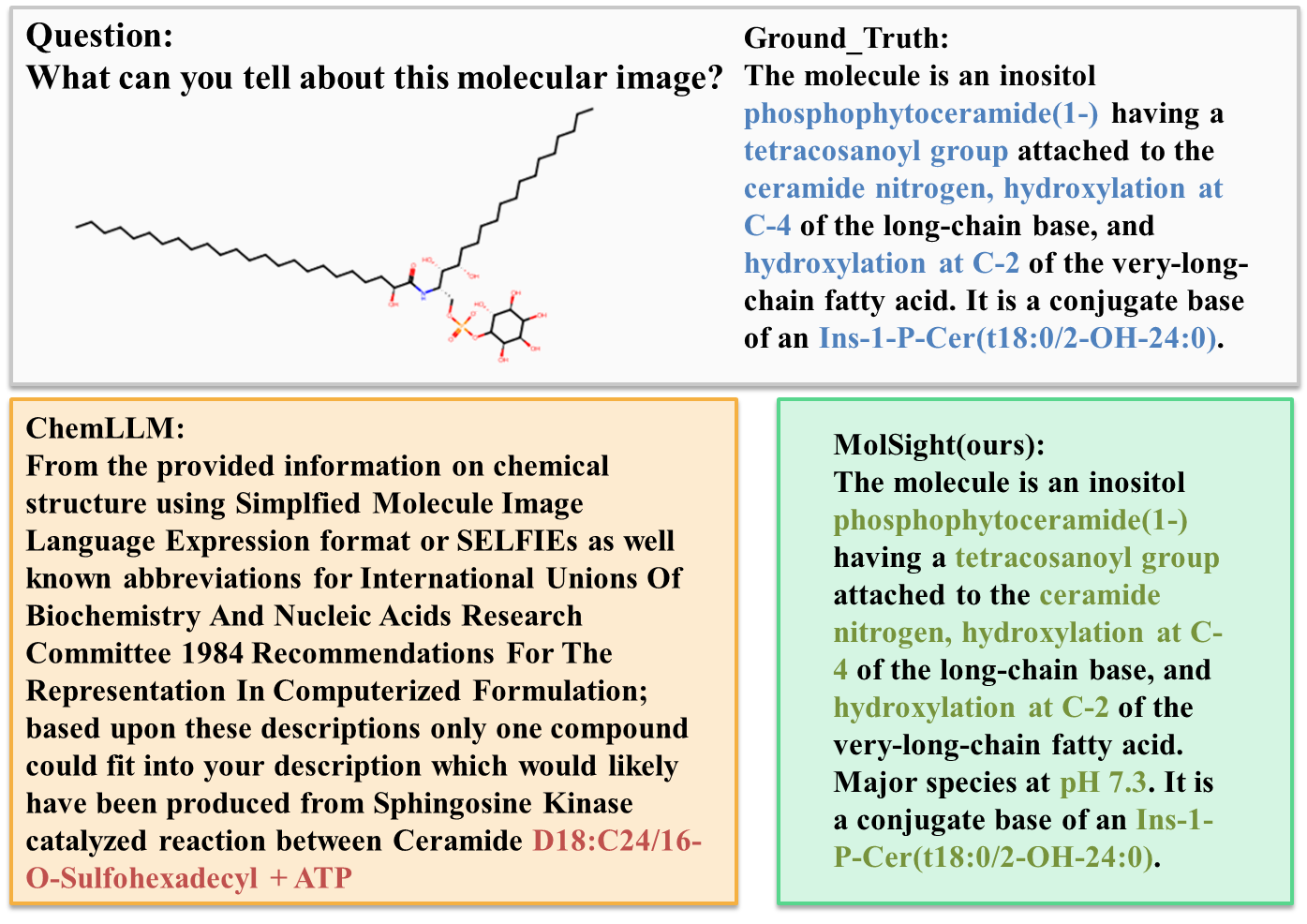}
\caption{Representative molecular captioning comparison between MolSight and ChemLLM.}
\label{fig:caption_case_chemllm}
\end{figure}

As shown in Figures~\ref{fig:caption_case_chemvlm} and~\ref{fig:caption_case_chemllm}, we present representative comparison cases on the CHEBI-20~\citep{chebi20} test set against existing molecular LLMs and VLMs. In the comparison with ChemVLM~\citep{li2025chemvlm}, MolSight more accurately captures key chemical attributes, including molecular identity, functional groups, and structural categories, demonstrating that its advantage is not limited to automatic metrics such as BLEU and ROUGE but also reflects more faithful chemical understanding. In the comparison with ChemLLM~\citep{zhang2024chemllm}, MolSight not only reproduces the information covered by the ground-truth caption, but also provides chemically meaningful descriptions beyond the reference, such as the correct major species at pH 7.3. By contrast, ChemLLM produces an incorrect synthetic-route description, indicating weaker grounding in the molecular image.

\phantomsection
\subsection*{E. Statistical Analysis}\label{app:statistical_analysis}\mbox{}\\

\phantomsection
\subsubsection*{E.1 Additional Analysis on Stereochemical Information}\label{app:stereo_analysis}\mbox{}\\

Molecular images may contain stereochemical cues, such as wedge bonds, dashed bonds, and directional double-bond annotations. Therefore, beyond the main image-to-SMILES metrics, we further evaluate whether MolSight can recover stereochemical information from molecular images. Since not all ground-truth molecules contain stereochemical annotations, we adopt an adaptive evaluation protocol that distinguishes stereochemical and non-stereochemical molecules.

Specifically, if the ground-truth molecule contains stereochemical annotations, a prediction is considered correct only when the stereochemistry is exactly recovered. Otherwise, if the ground truth does not contain stereochemical annotations, the prediction is considered correct as long as the underlying 2D molecular graph matches, regardless of whether the predicted SMILES contains stereochemical marks. Adaptive Sim follows the same principle: it applies stereochemistry-aware similarity only to molecules with stereochemical annotations and uses standard fingerprint similarity otherwise.

As shown in Table~\ref{tab:stereo_analysis}, MolSight achieves high Adaptive Sim and Adaptive ACC, suggesting that it can recover molecular structures with strong overall fidelity under this adaptive protocol. A more detailed comparison between Stereo ACC and Non-stereo ACC reveals the main source of remaining errors: Non-stereo ACC is already very high for both models and reaches 0.9892 for MolSight-32B, whereas Stereo ACC is substantially lower. Since Adaptive ACC aggregates both stereochemical and non-stereochemical subsets, this gap indicates that most remaining errors come from stereochemical recovery rather than ordinary 2D molecular graph recognition. In addition, MolSight-32B improves over MolSight-8B across all adaptive metrics, showing that larger backbone capacity further benefits stereochemical information recovery.

\begin{table*}[t]
\centering
\caption{Additional evaluation of stereochemical information learning in image-to-SMILES translation. Adaptive ACC and Adaptive Sim apply stereochemistry-aware evaluation only when the ground-truth molecule contains stereochemical annotations; otherwise, they evaluate whether the predicted 2D molecular graph matches the ground truth. Stereo ACC and Non-stereo ACC report results on molecules with and without stereochemical annotations, respectively.}
\label{tab:stereo_analysis}
\small
\setlength{\tabcolsep}{4pt}
\begin{tabular}{lcccc}
\toprule
Model & Adaptive ACC$\uparrow$ & Adaptive Sim$\uparrow$ & Stereo ACC$\uparrow$ & Non-stereo ACC$\uparrow$ \\
\midrule
MolSight-8B 
& 0.8344 $\pm$ 0.0043
& 0.9674 $\pm$ 0.0010 
& 0.4819 $\pm$ 0.0138 
& 0.9840 $\pm$ 0.0011 \\
MolSight-32B 
& \textbf{0.8519 $\pm$ 0.0022}
& \textbf{0.9703 $\pm$ 0.0007} 
& \textbf{0.5284 $\pm$ 0.0061} 
& \textbf{0.9892 $\pm$ 0.0006} \\
\bottomrule
\end{tabular}
\end{table*}

\phantomsection
\subsubsection*{E.2. SMILES Translation Analysis}\label{app:smiles_analysis}\mbox{}\\

\begin{figure}[!ht]
\centering
\includegraphics[width=0.5\textwidth]{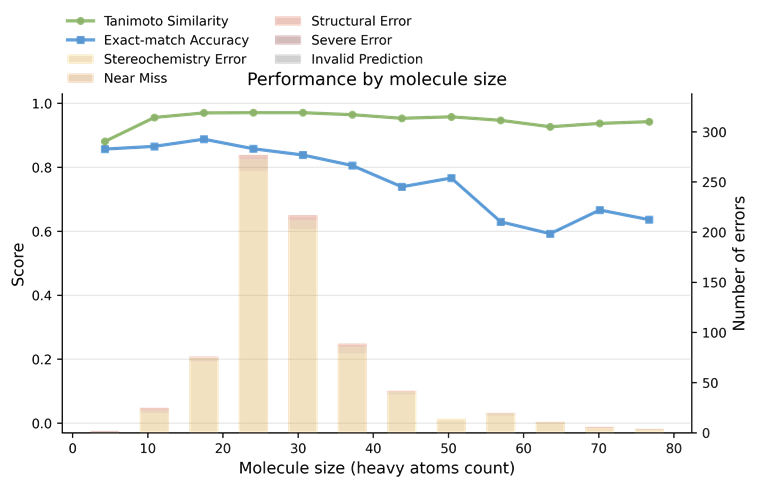}
\caption{Statistical analysis of MolSight on SMILES translation. The line plots show Tanimoto similarity and exact-match accuracy across different molecule-size bins, while the stacked bars show the distribution of error types in each bin.}
\label{fig:img2smiles_stat_analysis}
\end{figure}

For SMILES translation, we provide a supplementary analysis of MolSight from two perspectives: sensitivity to molecular length and error-type distribution, as shown in Figure~\ref{fig:img2smiles_stat_analysis}. The two line plots show how Tanimoto similarity and exact-match accuracy vary with molecular length. Here, molecular length is measured by the number of heavy atoms computed by RDKit when available, and falls back to SMILES length when RDKit parsing fails. The results show that Tanimoto similarity remains consistently high as molecule size increases, indicating that MolSight can stably recover the overall structural similarity even for longer and more complex molecules. In contrast, exact-match accuracy is more sensitive to molecule size and shows a decreasing trend as molecular length increases. This suggests that larger molecules, which contain more atoms, branches, ring systems, and stereochemical information, substantially increase the difficulty of exact structure recovery. Even when the predicted molecule remains structurally close to the ground truth, local differences can prevent an exact match. Nevertheless, the lowest accuracy across molecule-size bins still remains above 0.6, which is higher than the average performance of the strongest baseline.

The error analysis explains the source of this performance variation. We categorize prediction errors into four major types. Stereochemistry errors correspond to cases where the two-dimensional structure is nearly identical but the chiral or stereochemical annotations are inconsistent. Near misses indicate predictions with high structural similarity to the ground truth, defined by similarity $\geq 0.8$. Structural errors correspond to more evident structural deviations with $0.2 \leq$ similarity $< 0.8$, while severe errors denote predictions with similarity $< 0.2$. Invalid predictions refer to generated SMILES strings that cannot be parsed. Overall, among the 810 error samples in this run, stereochemistry-related errors account for the majority (726), while near misses (45), structural errors (21), severe errors (1), and invalid predictions (17) are relatively rare. This indicates that the dominant failure mode of MolSight lies in stereochemical or fine-grained local discrepancies rather than complete structural misunderstanding. 

\begin{figure}[!ht]
\centering
\includegraphics[width=\columnwidth]{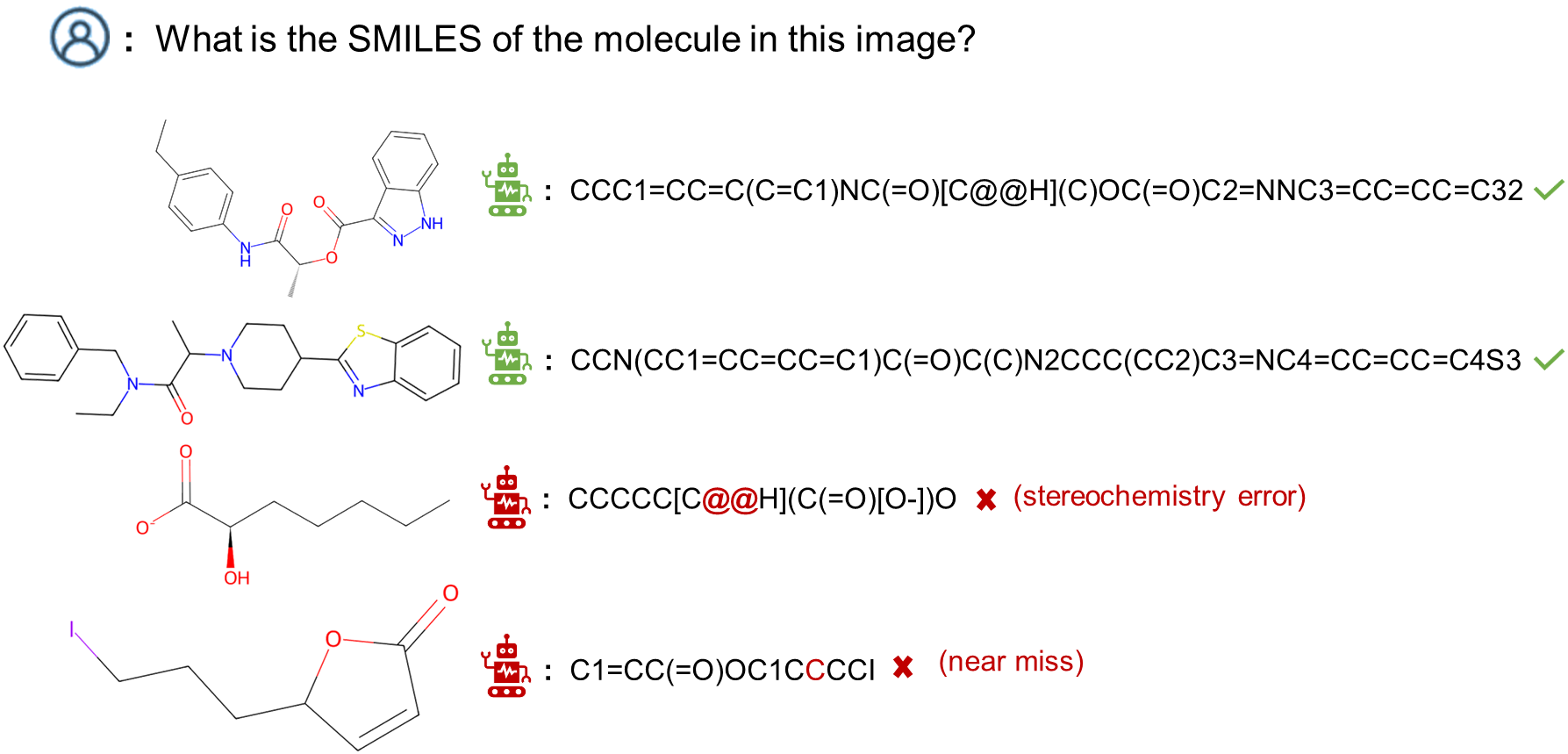}
\caption{
Representative examples for the SMILES translation error analysis.
Green check marks indicate exact-match predictions, while red crosses indicate typical error cases.
The last two examples illustrate the dominant failure modes identified in our statistical analysis, including stereochemistry errors and near-miss predictions, where the generated molecule remains largely similar to the ground truth but differs in stereochemical or local structural details.
}
\label{fig:img2smiles_error_examples}
\end{figure}

Figure~\ref{fig:img2smiles_error_examples} provides representative examples corresponding to the error categories discussed above. In the stereochemistry-error case, the prediction contains an extra ``@'' annotation; in the near-miss case, it introduces an additional carbon atom ``C'', causing a local structural mismatch. These results demonstrate that MolSight has strong overall structure perception ability, while also suggesting that exact recovery of complex molecules and learning image-implied three-dimensional stereochemical rules remain important directions for further improvement.

\phantomsection
\subsubsection*{E.3 Molecular Captioning Analysis}\label{app:captioning_analysis}\mbox{}\\

\begin{figure}[!ht]
\centering
\includegraphics[width=\columnwidth]{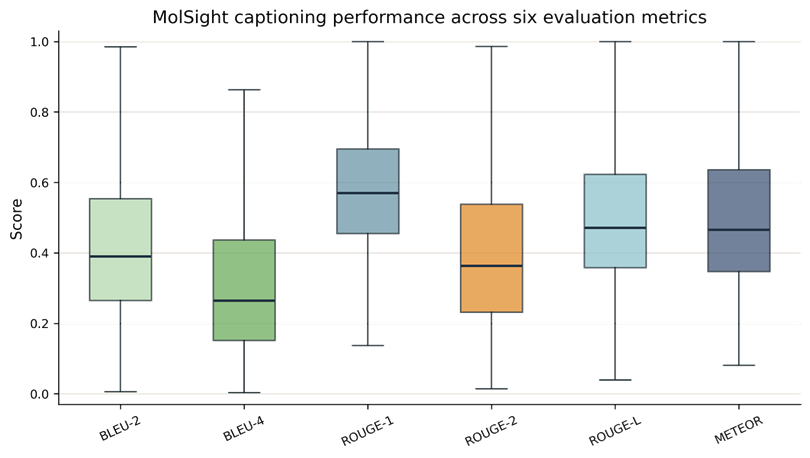}

\vspace{0.5em}

\includegraphics[width=\columnwidth]{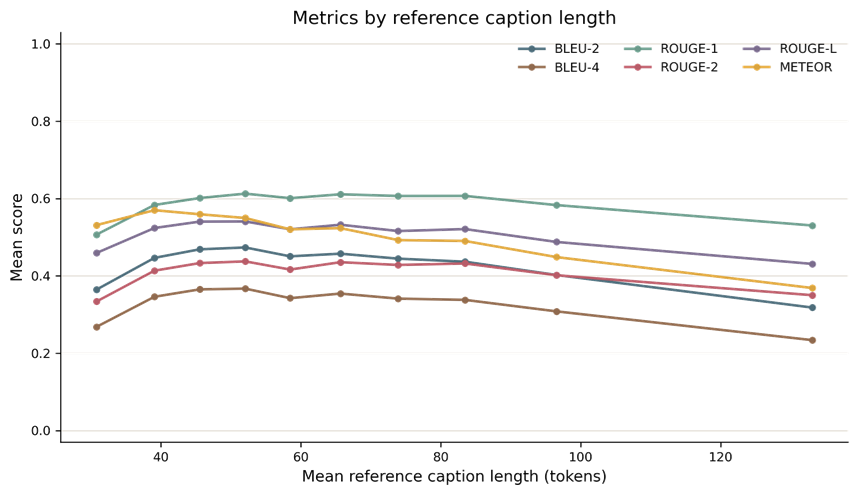}
\caption{
Statistical analysis of MolSight's molecular captioning performance.
Top: score distributions across six NLP-based evaluation metrics.
Bottom: metric trends across different reference-caption length.
}
\label{fig:captioning_statistical_analysis}
\end{figure}

For molecular captioning, since the MoleculeQA benchmark is formulated as a multiple-choice task, sample-level statistical analysis is less informative. Therefore, we further conduct statistical analysis based on the generated molecular captions. The first plot in Figure~\ref{fig:captioning_statistical_analysis} shows the distribution of MolSight's performance across NLP-based metrics. Overall, the six metrics exhibit relatively concentrated distributions, indicating that MolSight achieves stable molecular captioning performance across different samples without obvious long-tail degradation. The second plot further analyzes how these metrics vary with reference-caption length. The results show that most metrics gradually decrease as the reference captions become longer. This suggests that longer molecular descriptions usually contain more structural, property-related, and functional information, requiring the model to cover more fine-grained semantics and thus increasing the generation difficulty. Nevertheless, the overall trends are smooth and do not show clear performance collapse, indicating that MolSight maintains stable text generation ability across molecular descriptions of different lengths.

\phantomsection
\subsubsection*{E.4 Descriptor Estimation Analysis}\label{app:property_analysis}\mbox{}\\

\begin{figure*}[!ht]
\centering
\includegraphics[width=\textwidth]{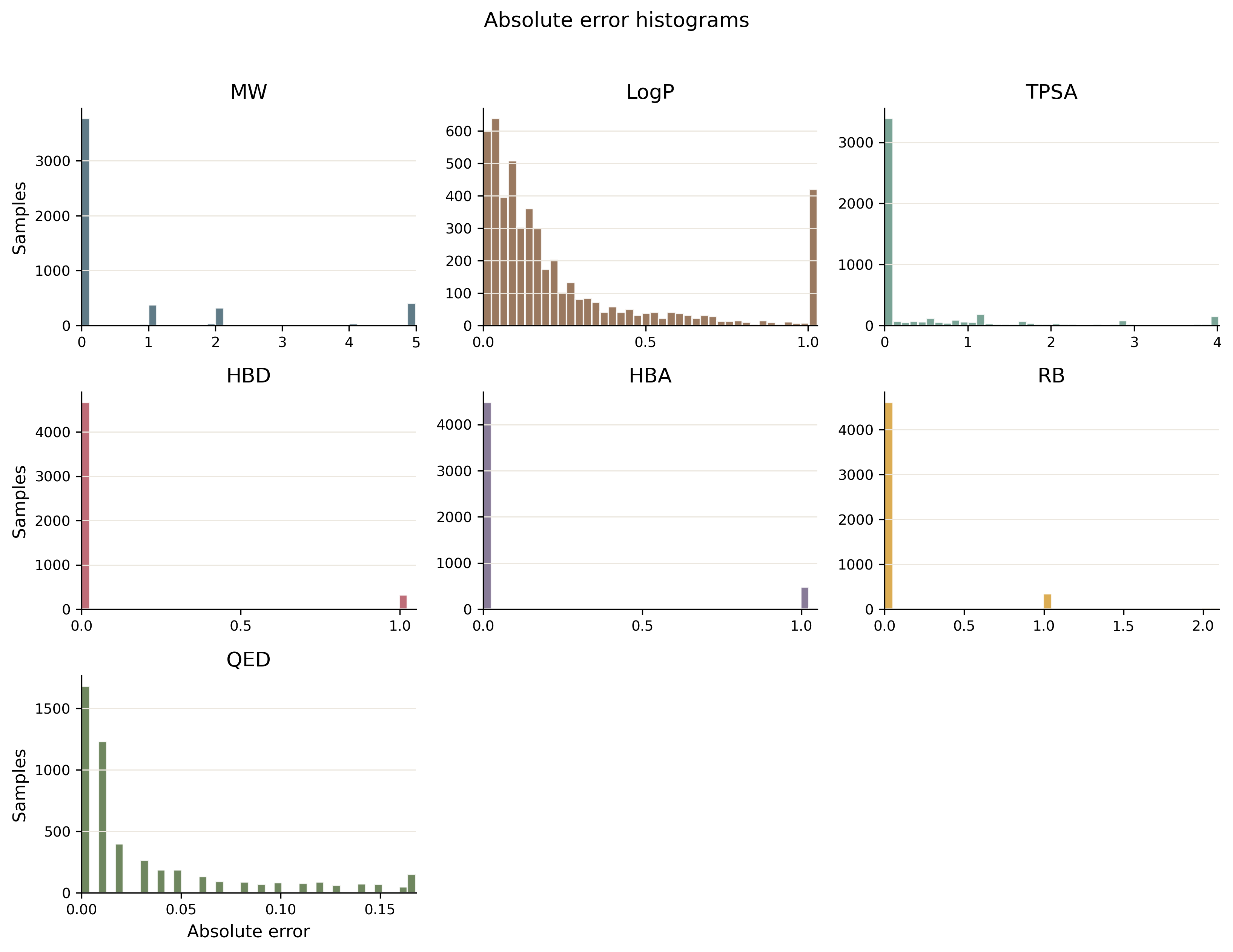}
\caption{
Absolute error distributions for physicochemical descriptor prediction.
Each subplot shows the distribution of absolute prediction errors for one descriptor.
}
\label{fig:property_abs_error_histograms}
\end{figure*}

\begin{figure*}[t]
\centering
\includegraphics[width=\textwidth]{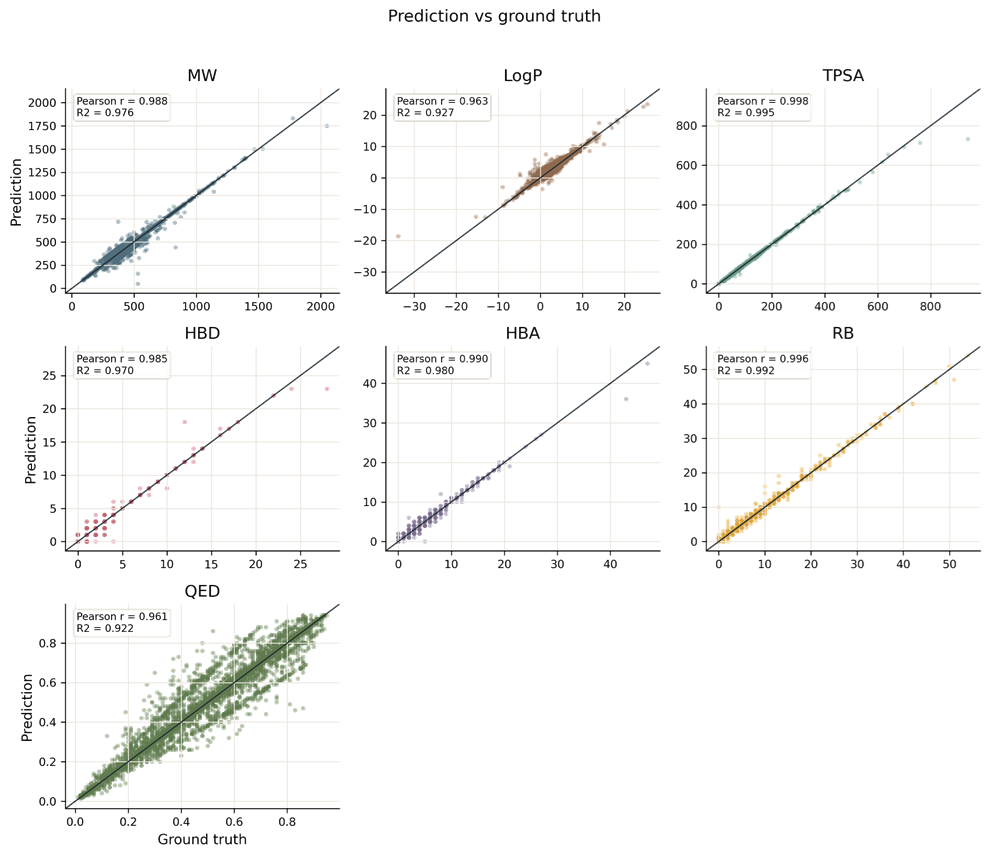}
\caption{
Prediction vs. ground-truth scatter plots for physicochemical descriptor prediction.
Each subplot reports the Pearson correlation coefficient and $R^2$ score, with the diagonal line indicating perfect prediction.
}
\label{fig:property_pred_vs_gt}
\end{figure*}

For descriptor estimation, we further analyze MolSight from two perspectives: error distribution and prediction correlation. As shown in Figure~\ref{fig:property_abs_error_histograms}, the absolute errors of the seven physicochemical descriptors are mostly concentrated near 0, indicating that the predicted values of most samples are very close to the ground truth and that the model maintains low errors across different types of numerical properties. Figure~\ref{fig:property_pred_vs_gt} further shows the scatter plots between predicted values and ground-truth values, where a strong correlation can be clearly observed for each property. These statistical results further demonstrate that MolSight not only achieves strong average performance but also performs stable and accurate molecular property estimation at the sample level.

\phantomsection
\subsubsection*{E.5 Out-of-Distribution Analysis}\label{app:ood_analysis}\mbox{}\\
To demonstrate the robustness of MolSight across different tasks, we conduct out-of-distribution analysis on both SMILES translation and descriptor estimation. When defining the distribution shift, we consider two complementary dimensions: molecular scaffold and fingerprint similarity.

\textbf{Scaffold-based OOD.}
This setting evaluates whether the model can generalize to molecules with unseen core structures, reflecting a scaffold-shift scenario beyond memorizing familiar molecular backbones.
Specifically, we extract the Bemis--Murcko scaffold for each molecule in the training and test sets, and regard test samples whose scaffolds do not appear in the training set as OOD samples.

\textbf{Fingerprint-similarity-based OOD.}
This setting measures how far a test molecule is from the training distribution in terms of overall molecular structure, including substituents, functional groups, and fine-grained local structural patterns.
For each test molecule, we compute its maximum Tanimoto similarity to all training molecules using Morgan fingerprints, and treat the bottom 25\% of test samples with the lowest maximum train-test similarity as OOD samples.

\begin{table}[!ht]
\centering
\caption{OOD analysis on SMILES translation. ID denotes in-distribution samples, while OOD denotes out-of-distribution samples.}
\label{tab:ood_smiles}
\small
\setlength{\tabcolsep}{6pt}
\resizebox{\columnwidth}{!}{
\begin{tabular}{llccc}
\toprule
OOD Setting & Split & Avg. Sim$\uparrow$ & ACC$\uparrow$ & Valid$\%\uparrow$ \\
\midrule
\multirow{3}{*}{Scaffold}
& All & 0.967 & 0.841 & 99.7\% \\
& ID (54\%) & 0.969 & 0.850 & 100\% \\
& OOD (46\%) & 0.966 & 0.829 & 99.5\% \\
\midrule
\multirow{3}{*}{Fingerprint}
& All & 0.967 & 0.841 & 99.7\% \\
& ID (75\%) & 0.971 & 0.849 & 99.8\% \\
& OOD (25\%) & 0.958 & 0.815 & 99.4\% \\
\bottomrule
\end{tabular}
}
\end{table}

\begin{table*}[!ht]
\centering
\caption{OOD analysis on descriptor estimation. ID denotes in-distribution samples, while OOD denotes out-of-distribution samples.}
\label{tab:ood_descriptor}
\small
\setlength{\tabcolsep}{3pt}
\resizebox{\textwidth}{!}{
\begin{tabular}{llcccccccccccccccc}
\toprule
\multirow{2}{*}{OOD Setting} & \multirow{2}{*}{Split}
& \multicolumn{2}{c}{MW}
& \multicolumn{2}{c}{LogP}
& \multicolumn{2}{c}{TPSA}
& \multicolumn{2}{c}{HBD}
& \multicolumn{2}{c}{HBA}
& \multicolumn{2}{c}{RB}
& \multicolumn{2}{c}{QED}
& \multicolumn{2}{c}{Avg.} \\
\cmidrule(lr){3-4}
\cmidrule(lr){5-6}
\cmidrule(lr){7-8}
\cmidrule(lr){9-10}
\cmidrule(lr){11-12}
\cmidrule(lr){13-14}
\cmidrule(lr){15-16}
\cmidrule(lr){17-18}
& & Pearson$\uparrow$ & MAE$\downarrow$
& Pearson$\uparrow$ & MAE$\downarrow$
& Pearson$\uparrow$ & MAE$\downarrow$
& Pearson$\uparrow$ & MAE$\downarrow$
& Pearson$\uparrow$ & MAE$\downarrow$
& Pearson$\uparrow$ & MAE$\downarrow$
& Pearson$\uparrow$ & MAE$\downarrow$
& Pearson$\uparrow$ & MAE$\downarrow$ \\
\midrule
\multirow{3}{*}{Scaffold}
& All
& 0.995 & 1.777
& 0.974 & 0.301
& 0.999 & 0.982
& 0.996 & 0.014
& 0.992 & 0.091
& 0.991 & 0.248
& 0.943 & 0.047
& 0.984 & 0.494 \\
& ID (54\%)
& 0.997 & 1.189
& 0.978 & 0.268
& 0.999 & 0.752
& 0.998 & 0.009
& 0.992 & 0.066
& 0.992 & 0.204
& 0.935 & 0.047
& 0.984 & 0.354 \\
& OOD (46\%)
& 0.995 & 2.528
& 0.973 & 0.341
& 0.998 & 1.252
& 0.995 & 0.021
& 0.991 & 0.121
& 0.990 & 0.300
& 0.950 & 0.046
& 0.985 & 0.659 \\
\midrule
\multirow{3}{*}{Fingerprint}
& All
& 0.995 & 1.777
& 0.974 & 0.301
& 0.999 & 0.982
& 0.996 & 0.014
& 0.992 & 0.091
& 0.991 & 0.248
& 0.943 & 0.047
& 0.984 & 0.494 \\
& ID (75\%)
& 0.999 & 0.900
& 0.978 & 0.276
& 0.999 & 0.906
& 0.998 & 0.013
& 0.993 & 0.073
& 0.992 & 0.222
& 0.950 & 0.044
& 0.987 & 0.348 \\
& OOD (25\%)
& 0.989 & 4.405
& 0.968 & 0.378
& 0.998 & 1.210
& 0.992 & 0.018
& 0.989 & 0.146
& 0.986 & 0.327
& 0.921 & 0.057
& 0.978 & 0.934 \\
\bottomrule
\end{tabular}
}
\end{table*}

Tables~\ref{tab:ood_smiles}-\ref{tab:ood_descriptor} show that MolSight maintains strong molecular structure recovery and property prediction ability on molecules with unseen scaffolds or low overall structural similarity to the training set. This suggests that the model does not simply memorize training molecules, but instead learns a robust visual--structural mapping.

\phantomsection
\subsection*{F. Computational Efficiency}\label{app:computational_efficiency}\mbox{}\\

Table~\ref{tab:computational_efficiency} reports the computational resources, training configuration settings, and training time used at each stage of MolSight.
\begin{table*}[!ht]
\centering
\caption{Computational efficiency and training hyperparameters of MolSight.}
\label{tab:computational_efficiency}
\small
\setlength{\tabcolsep}{4pt}
\resizebox{\textwidth}{!}{
\begin{tabular}{lccccccc}
\toprule
\multirow{2}{*}{Setting}
& \multirow{2}{*}{Substage 1}
& \multirow{2}{*}{Substage 2}
& \multicolumn{5}{c}{Stage 2} \\
\cmidrule(lr){4-8}
& & 
& SMILES translation
& Captioning (MoleculeQA)
& Captioning (CHEBI-20)
& Descriptor estimation
& Bioactivity prediction \\
\midrule
Base Model 
& \multicolumn{7}{c}{Qwen3-VL-8B-Instruct} \\
GPUs
& \multicolumn{7}{c}{NVIDIA A800/A100 80GB} \\
\#GPUs
& 4 & 4 & 2 & 2 & 2 & 2 & 2 \\
Trainable Params
& 407.6M & 201.4M & 90.4M & 90.4M & 90.4M & 90.4M & 180.9M \\
Epochs
& 0.5 & 0.5 & 2 & 2 & 2 & 2 & 2 \\
Training Steps 
& 2400 & 2400 & 5938 & 4689 & 4352 & 5938 & - \\
Training Hours
& $\sim$50h & $\sim$33h & $\sim$18h & $\sim$18h & $\sim$19h & $\sim$19h & $\sim$10h (all 6 tasks) \\
Batch Size / GPU
& 2 & 2 & 2 & 2 & 2 & 2 & 2 \\
Grad. Accum.
& 16 & 16 & 8 & 8 & 8 & 8 & 8 \\
Learning Rate
& $1\times10^{-4}$ & $5\times10^{-5}$ & $2\times10^{-4}$ & $2\times10^{-4}$ & $2\times10^{-4}$ & $2\times10^{-4}$ & $2\times10^{-4}$ \\
Optimizer
& \multicolumn{7}{c}{AdamW} \\
Warmup Ratio
& \multicolumn{7}{c}{0.03} \\
Weight Decay
& \multicolumn{7}{c}{0.01} \\
Grad. Clip
& \multicolumn{7}{c}{1.0} \\
\bottomrule
\end{tabular}
}
\end{table*}

\phantomsection
\subsection*{G. Additional Ablation Results}\label{app:additional_ablation}\mbox{}\\
\textbf{Effect of molecular images.}
To verify that molecular images provide substantial benefits for molecular understanding, we conduct an image-input ablation study on the SMILES translation task. Since our topology adaptation layer requires SVG input, we evaluate the effect of molecular images during Stage 2 fine-tuning. As shown in Table~\ref{tab:ablation_image_input}, adding the molecular image consistently improves all metrics compared with the variant without image input. This result indicates that molecular images contribute complementary visual structural information beyond SVG annotations, leading to consistent improvements in molecular structure recovery.
\begin{table}[!ht]
\centering
\caption{Ablation study on the effect of molecular image input in Stage 2 fine-tuning for SMILES translation. Avg. Sim denotes Tanimoto similarity. Higher is better for all metrics.}
\label{tab:ablation_image_input}
\small
\setlength{\tabcolsep}{6pt}
\begin{tabular}{lccc}
\toprule
Setting & Avg. Sim$\uparrow$ & ACC$\uparrow$ & Valid (\%)$\uparrow$ \\
\midrule
Direct fine-tune 
& 0.931 & 0.500 & 85.8\% \\
Direct fine-tune w/o image 
& 0.928 & 0.460 & 82.3\% \\
\bottomrule
\end{tabular}
\end{table}

\textbf{LoRA hyperparameter sensitivity.}
We further present a comprehensive overview of the ablation studies on LoRA hyperparameters in Table~\ref{tab:ablation_lora}. We report the LoRA hyperparameter ablation results in Tables~\ref{tab:lora_smiles_translation}--\ref{tab:lora_molecular_captioning_ablation}. Specifically, we compare different LoRA configurations, including the rank $r$ and scaling factor $\alpha$, across SMILES translation, the MoleculeQA benchmark, descriptor estimation, and molecular caption generation. The r defines the rank of the LoRA adapter matrices A and B, controlling the amount of information that the adapter can learn and express. $\alpha$ is used to adjust the amplitude of the output of the LoRA adapter. Both jointly control the size of the LoRA fine-tuning trainable parameters.

These results show that MolSight performs best on most (3/4) experiments under the set of hyperparameters r=32 and $\alpha$=64. It can be seen that after our MolSight perceiving and learning the molecular graph structure through the topology adaptation layer, only after lightweight fine-tune, it can be adapted to various molecular understanding tasks.

\begin{table}[H]
\centering
\caption{LoRA hyperparameter ablation on SMILES translation.}
\label{tab:lora_smiles_translation}
\begin{tabular}{lccc}
\toprule
Setting & Avg. Sim $\uparrow$ & ACC $\uparrow$ & Valid(\%) $\uparrow$ \\
\midrule
$r=64,\ \alpha=128$ & \textbf{0.998} & 0.805 & \textbf{99.7\%} \\
\rowcolor{blue!8}
$r=32,\ \alpha=64$  & \textbf{0.998} & \textbf{0.807} & \textbf{99.7\%} \\
$r=16,\ \alpha=32$  & 0.997 & 0.791 & 99.5\% \\
\bottomrule
\end{tabular}
\end{table}

\begin{table*}[!t]
\centering
\caption{LoRA hyperparameter ablation on MoleculeQA captioning benchmark.}
\label{tab:lora_moleculeqa_ablation}
\begin{tabular}{lccccc}
\toprule
Setting & Structure & Source & Property & Application & Total \\
\midrule
\rowcolor{blue!8}
$r=32,\ \alpha=64$  & \textbf{78.38} & \textbf{73.42} & \textbf{51.16} & \textbf{50.42} & \textbf{70.90} \\
$r=64,\ \alpha=128$ & 70.58 & 71.33 & 49.38 & 48.08 & 65.74 \\
\bottomrule
\end{tabular}
\end{table*}

\begin{table*}[!t]
\centering
\caption{LoRA hyperparameter ablation on descriptor estimation.}
\label{tab:lora_ablation_descriptor}
\small
\setlength{\tabcolsep}{3pt}
\resizebox{\textwidth}{!}{
\begin{tabular}{lcccccccccccccccc}
\toprule
\multirow{2}{*}{Method}
& \multicolumn{2}{c}{MW}
& \multicolumn{2}{c}{LogP}
& \multicolumn{2}{c}{TPSA}
& \multicolumn{2}{c}{HBD}
& \multicolumn{2}{c}{HBA}
& \multicolumn{2}{c}{RB}
& \multicolumn{2}{c}{QED}
& \multicolumn{2}{c}{Avg.} \\
\cmidrule(lr){2-3}
\cmidrule(lr){4-5}
\cmidrule(lr){6-7}
\cmidrule(lr){8-9}
\cmidrule(lr){10-11}
\cmidrule(lr){12-13}
\cmidrule(lr){14-15}
\cmidrule(lr){16-17}
& Pearson$\uparrow$ & MAE$\downarrow$
& Pearson$\uparrow$ & MAE$\downarrow$
& Pearson$\uparrow$ & MAE$\downarrow$
& Pearson$\uparrow$ & MAE$\downarrow$
& Pearson$\uparrow$ & MAE$\downarrow$
& Pearson$\uparrow$ & MAE$\downarrow$
& Pearson$\uparrow$ & MAE$\downarrow$
& Pearson$\uparrow$ & MAE$\downarrow$ \\
\midrule
MolSight $(r=64,\alpha=128)$
& 0.988 & 5.317
& 0.963 & 0.303
& 0.998 & \textbf{0.661}
& 0.985 & 0.071
& 0.990 & 0.112
& \textbf{0.996} & \textbf{0.092}
& \textbf{0.961} & \textbf{0.034}
& 0.983 & 0.941 \\
\rowcolor{blue!8}
\textbf{MolSight $(r=32,\alpha=64)$}
& \textbf{0.996} & \textbf{1.777}
& \textbf{0.975} & \textbf{0.301}
& \textbf{0.999} & 0.982
& 0.996 & 0.014
& \textbf{0.992} & \textbf{0.091}
& 0.991 & 0.248
& 0.943 & 0.047
& \textbf{0.984} & \textbf{0.494} \\
MolSight $(r=16,\alpha=32)$
& 0.995 & 1.905
& 0.969 & \textbf{0.301}
& 0.998 & 1.021
& \textbf{0.997} & \textbf{0.013}
& 0.989 & 0.100
& 0.989 & 0.264
& 0.941 & 0.047
& 0.982 & 0.521 \\
\bottomrule
\end{tabular}
}
\end{table*}

\begin{table*}[!t]
\centering
\caption{LoRA hyperparameter ablation on caption generation.}
\label{tab:lora_molecular_captioning_ablation}
\begin{tabular}{lcccccc}
\toprule
Setting & BLEU-2 $\uparrow$ & BLEU-4 $\uparrow$ & ROUGE-1 $\uparrow$ & ROUGE-2 $\uparrow$ & ROUGE-L $\uparrow$ & METEOR $\uparrow$ \\
\midrule
\rowcolor{blue!8}
$r=64,\ \alpha=128$ & \textbf{0.46} & \textbf{0.34} & \textbf{0.56} & \textbf{0.39} & \textbf{0.49} & \textbf{0.46} \\
$r=32,\ \alpha=64$  & 0.45 & 0.33 & 0.55 & 0.37 & 0.48 & \textbf{0.46} \\
\bottomrule
\end{tabular}
\end{table*}

\begin{table}[H]
\centering
\caption{Additional ablation results on attempted modules and auxiliary losses for SMILES translation. Avg. Sim denotes Tanimoto similarity. Higher is better for all metrics.}
\label{tab:additional_modules_losses}
\small
\setlength{\tabcolsep}{4pt}
\resizebox{\columnwidth}{!}{
\begin{tabular}{lccc}
\toprule
Setting & Avg. Sim$\uparrow$ & ACC$\uparrow$ & Valid (\%)$\uparrow$ \\
\midrule
\rowcolor{blue!8}
\textbf{MolSight} 
& \textbf{0.998} 
& \textbf{0.807} 
& 99.7\% \\
MolSight w/ token selector 
& 0.961 
& 0.680
& 99.3\% \\
MolSight w/ reconstruction loss 
& 0.980 
& 0.756 
& \textbf{99.8\%} \\
\bottomrule
\end{tabular}
}
\end{table}
\textbf{Additional modules and loss designs.}
We also experimented with several additional modules and auxiliary losses beyond the final MolSight design.
As shown in Table~\ref{tab:additional_modules_losses}, these additional designs do not improve over the final MolSight architecture.
First, we introduced a token selector module to address the sparsity of molecular images by selecting informative vision tokens before subsequent topology-aware processing.
However, this design does not work well in practice.
We hypothesize that the merger module in Qwen3-VL already plays a similar role in filtering and compressing vision tokens, so an additional token selection step may over-filter the visual representation and cause information loss.
Second, we added a reconstruction loss during training to reconstruct vision tokens as an auxiliary objective, but it also fails to bring consistent performance gains.
These results suggest that the final MolSight design provides a better balance between preserving visual structural information and injecting molecular topology.



\phantomsection

\phantomsection
\subsection*{H. Case Studies}\label{app:case_studies}\mbox{}\\

\phantomsection
\subsubsection*{H.1 SMILES Translation}\label{app:smiles_translation}
\begin{figure*}[!ht]
\centering
\includegraphics[width=\textwidth]{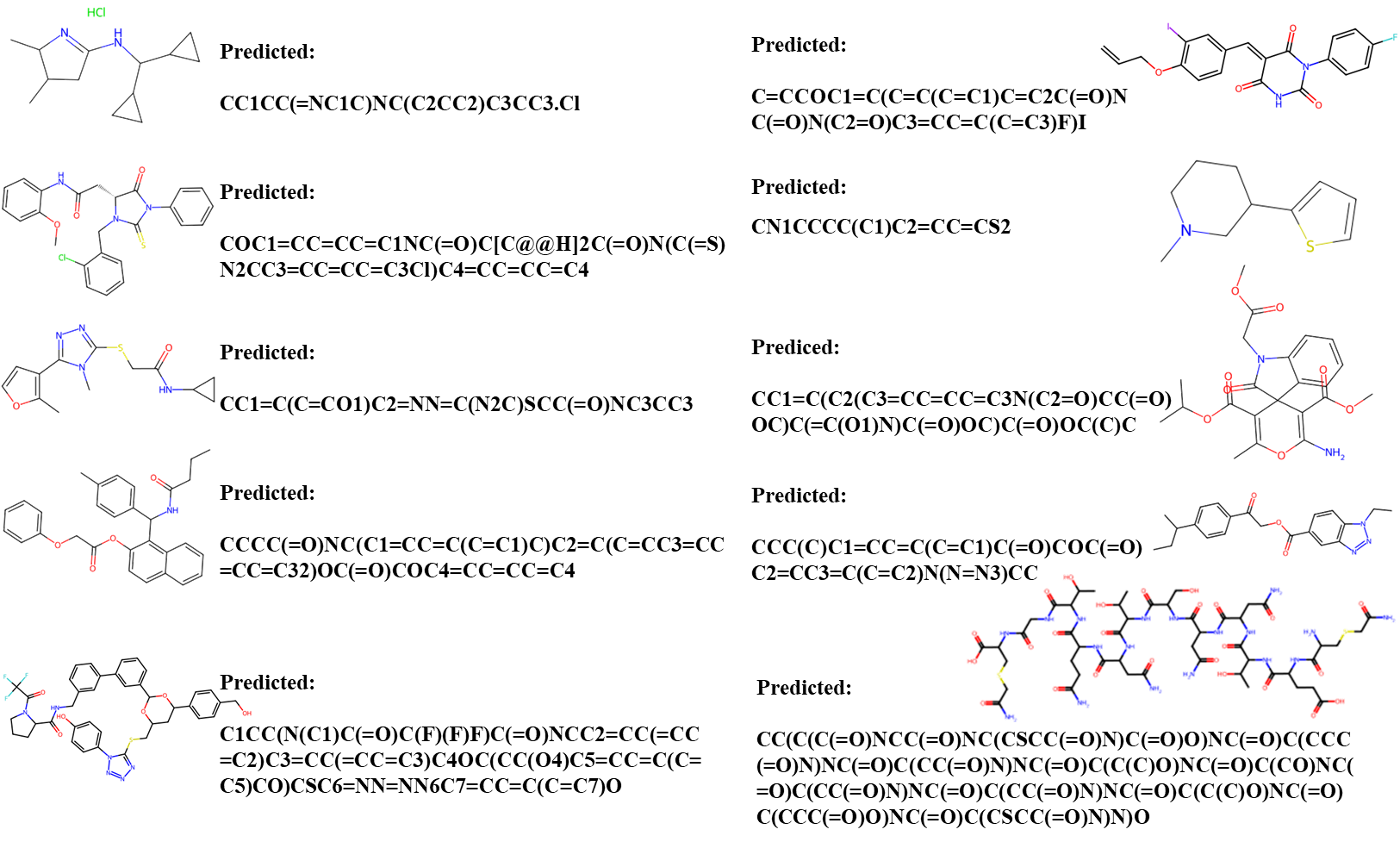}
\caption{Representative cases of MolSight on SMILES translation tasks.}
\label{fig:cases_smiles}
\end{figure*}

\mbox{}\\
As shown by the representative cases in Figure~\ref{fig:cases_smiles}, MolSight can accurately perform SMILES translation across diverse molecular visual scenarios, including general molecules, polycyclic systems, hydrochloride salts with external chemical compound, molecules with overlapping functional groups in the image, and highly large and complex molecular structures. These examples demonstrate the robustness of MolSight in recovering molecular structures from challenging and heterogeneous visual inputs.

\phantomsection
\subsubsection*{H.2 Molecular Captioning}\label{app:molecular_captioning}\mbox{}\\

\begin{figure*}[!ht]
\centering
\includegraphics[width=\textwidth]{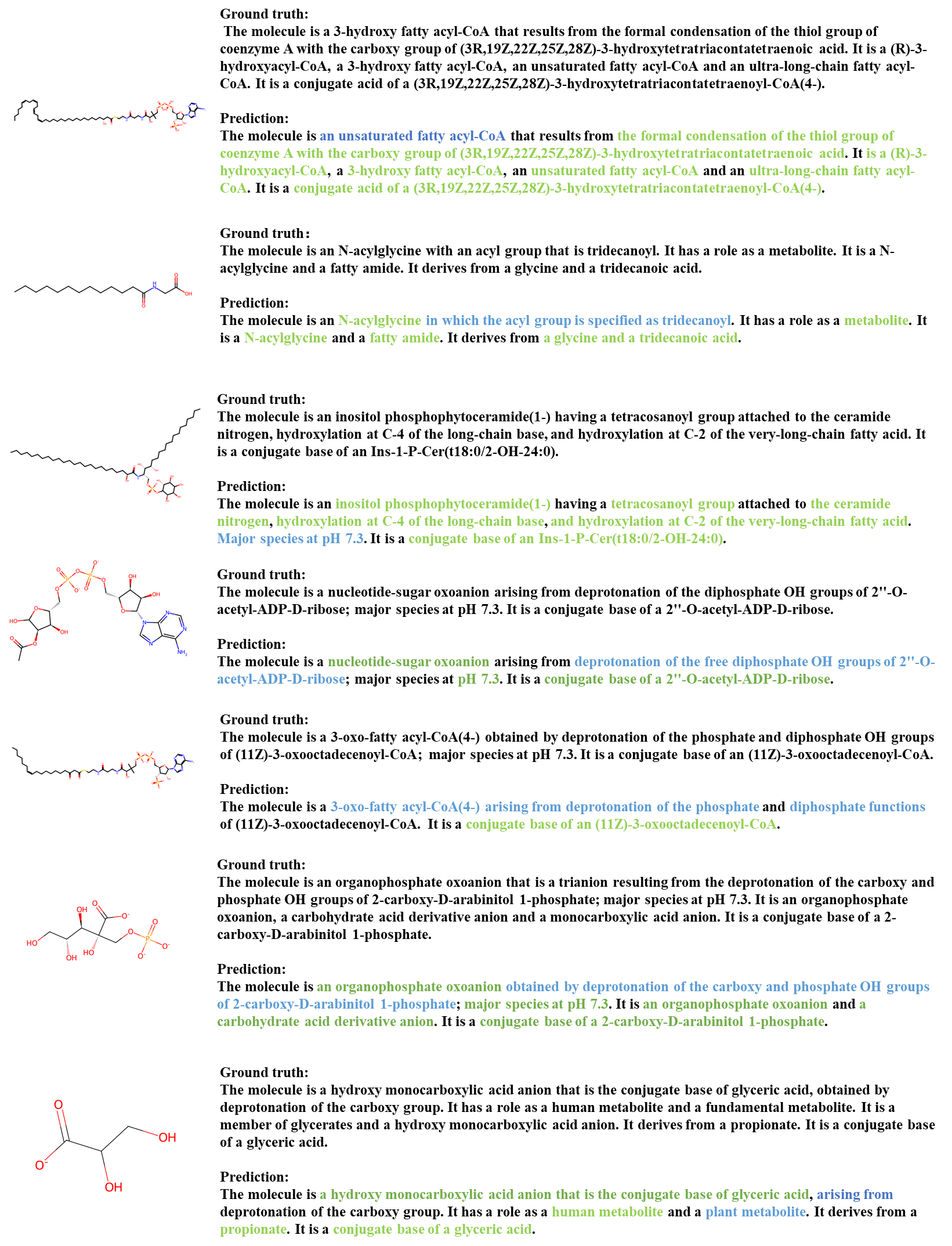}
\caption{Representative cases of MolSight on Molecular Captioning tasks. The green values indicate predictions strongly correlated with the ground truth, while the blue values indicate predictions that differ from or are not included in the ground truth but are verified as correct.}
\label{fig:cases_captioning}
\end{figure*}

As shown by the representative cases in Figure~\ref{fig:cases_captioning}, MolSight effectively learns the writing style of molecular descriptions and improves the accuracy of key information coverage, including molecular structure, function, physicochemical context, and synthetic accessibility.

\phantomsection
\subsubsection*{H.3 Descriptor Estimation}\label{app:descriptor_estimation}
\begin{figure*}[!ht]
\centering
\includegraphics[width=\textwidth]{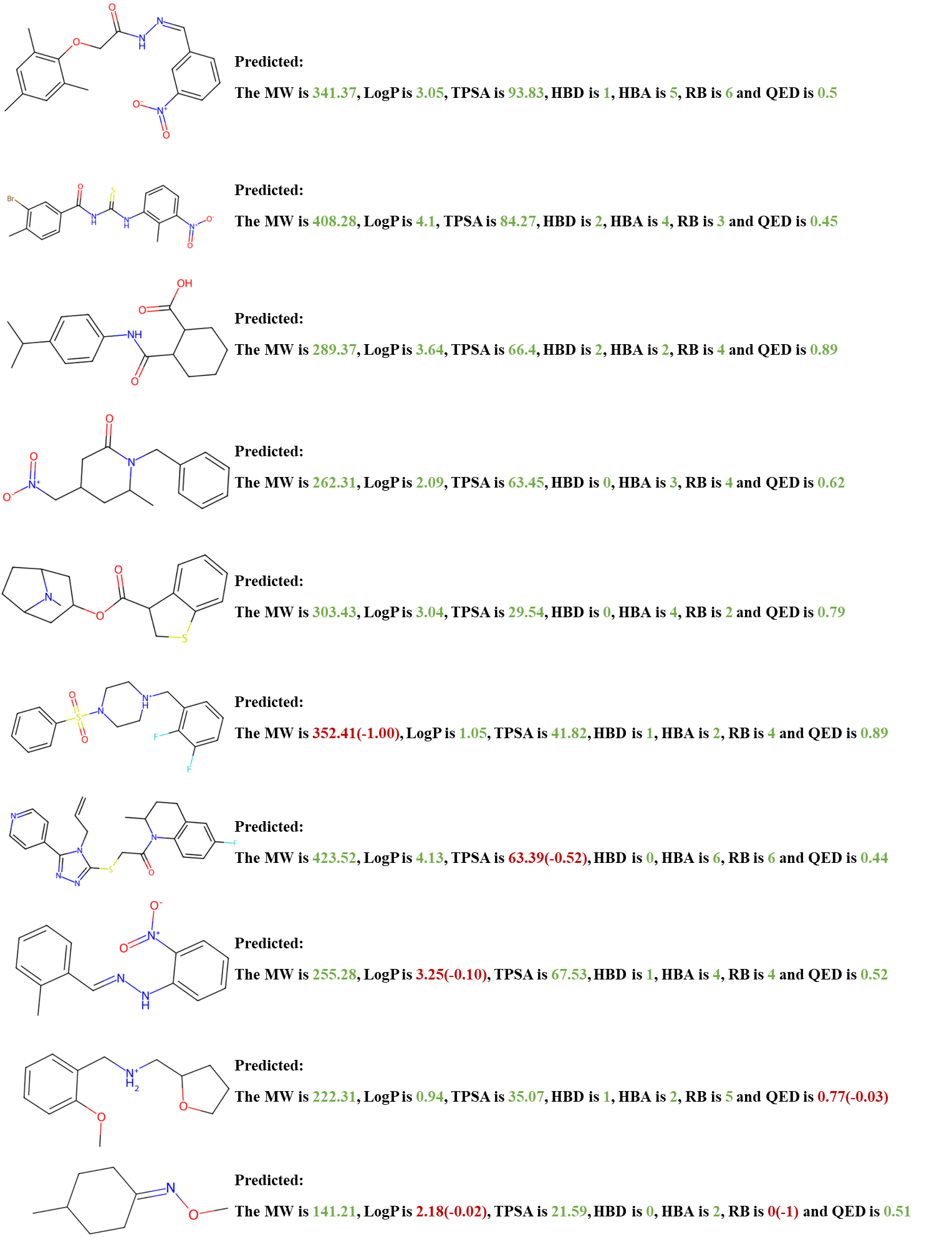}
\caption{Representative cases of MolSight on descriptor estimation tasks. The green values indicate correct predictions, while the red values indicate the difference between predicted and ground truth.}
\label{fig:cases_property}
\end{figure*}

\mbox{}\\
As shown by the representative cases in Figure~\ref{fig:cases_property}, MolSight produces descriptor estimates that are highly close to the ground truth across diverse molecular structures. Most examples show only minor deviations in one or a few properties, while a smaller subset achieves exact matches across all seven descriptors. The observed errors are generally property-specific and limited in magnitude, such as slight deviations in molecular weight, lipophilicity, polarity-related descriptors, QED, or rotatable bond count. These cases suggest that MolSight can reliably capture both global molecular composition and fine-grained structural cues from molecular images, leading to stable and near-accurate descriptor estimation.

\end{document}